%% file: neurips_2020.tex
\pdfoutput=1

\documentclass{article}

\usepackage[preprint]{neurips_2020}

\usepackage[utf8]{inputenc} 
\usepackage[T1]{fontenc}    
\usepackage{hyperref}       
\usepackage{url}            
\usepackage{booktabs}       
\usepackage{amsfonts}       
\usepackage{nicefrac}       
\usepackage{microtype}      
\usepackage{multicol}

\usepackage{xcolor}

\usepackage{algorithm, algorithmic}
\usepackage{caption, enumitem, bm, mathtools, bbm}
\usepackage{xspace}
\usepackage{subcaption}
\usepackage{multirow}

\newtheorem{thm}{Theorem}[section] 
\newtheorem{cor}{Corollary}[section]
\newtheorem{prop}{Proposition}[section]
\newtheorem{claim}{Claim}[section]
\newtheorem{definition}{Definition}[section]
\newtheorem{remark}{Remark}

\def\arginf{\mathop{\hbox{\rm arginf }}}
\def\argsup{\mathop{\hbox{\rm argsup }}}

\DeclareMathOperator{\kmax}{\text{kmax}}

\newcommand{\ourtoplayer}{\texttt{PLLay}\xspace}

\title{PLLay: Efficient Topological Layer \\ based on Persistence Landscapes}

\author{%
Kwangho Kim
    \\
  Carnegie Mellon University\\
  Pittsburgh, USA \\
  \texttt{kwanghk@cmu.edu} \\
  \And
  Jisu Kim \\
  Inria\\
  Palaiseau, France \\
  \texttt{jisu.kim@inria.fr} \\
  \AND
  Manzil Zaheer \\
  Google Research\\
  Mountain View, USA \\
  \texttt{manzilzaheer@google.com} \\
  \And
  Joon Sik Kim \\
  Carnegie Mellon University\\
  Pittsburgh, USA \\
  \texttt{joonsikk@cs.cmu.edu} \\
  \AND
  Frederic Chazal \\
  Inria\\
  Palaiseau, France \\ 
  \texttt{frederic.chazal@inria.fr} \\
  \And
  Larry Wasserman \\
  Carnegie Mellon University\\
  Pittsburgh, USA \\
  \texttt{larry@stat.cmu.edu} \\
}

\begin{document}

\maketitle

\input{toplayer_main.tex}


\bibliography{toplayer}
\bibliographystyle{plainnat}

\clearpage
\newpage

\appendix 

\input{toplayer_appendix.tex}

\end{document}

%% file: toplayer_main.tex
\begin{abstract}
    We propose \ourtoplayer, a novel topological layer for general deep learning models based on persistence landscapes, in which we can efficiently exploit the underlying topological features of the input data structure. In this work, we show differentiability with respect to layer inputs, for a general persistent homology with arbitrary filtration. Thus, our proposed layer can be placed anywhere in the network and feed critical information on the topological features of input data into subsequent layers to improve the learnability of the networks toward a given task. A task-optimal structure of \ourtoplayer is learned during training via backpropagation, without requiring any input featurization or data preprocessing. We provide a novel adaptation for the DTM function-based filtration, and show that the proposed layer is robust against noise and outliers through a stability analysis. We demonstrate the effectiveness of our approach by classification experiments on various datasets.
\end{abstract} 

\section{Introduction}
With its strong generalizability, deep learning has been pervasively applied in machine learning. To improve the learnability of deep learning models, various techniques have been proposed. Some of them have achieved an efficient data processing method through specialized layer structures; for instance, 
inserting a convolutional layer greatly improves visual object recognition and other tasks in computer vision \citep[e.g.,][]{krizhevsky2012imagenet, lecun2016lenet}. 
On the other hand, a large body of recent work focuses on optimal architecture of deep network \citep{simonyan2014very, he2016deep, szegedy2015going, albelwi2016automated}. 

In this paper, we explore an alternative way to enhance the learnability of deep learning models by developing a novel \textit{topological layer} which feeds the significant topological features of the underlying data structure in an arbitrary network. The power of topology lies in its capacity which differentiates sets in topological spaces in a robust and meaningful geometric way \citep{carlsson2009topology, ghrist2008barcodes}. It provides important insights into the global "shape" of the data structure via \textit{persistent homology} \citep{zomorodian2005computing}. The use of topological methods in data analysis has been limited by the difficulty of combining the main tool of the subject, {persistent homology}, with statistics and machine learning. Nonetheless, a series of recent studies have reported notable successes in utilizing topological methods in data analysis  \citep[e.g.,][]{zhu2013persistent, dindin2020topological, nanda2014simplicial, tralie2018quasi, seversky2016time, gamble2010exploring, pereira2015persistent, umeda2017time, liu2016applying, venkataraman2016persistent, emrani2014persistent}

There are at least three benefits of utilizing the topological layer in deep learning; 1) we can efficiently extract robust global features of input data that otherwise would not be readily accessible via traditional feature maps, 2) an optimal structure of the layer for a given task can be easily embodied via backpropagation during training, and 3) with proper filtrations it can be applied to arbitrarily complicated data structure even without any data preprocessing.

\textbf{Related Work.}
The idea of incorporating topological concepts into deep learning has been explored only recently, mostly via feature engineering perspective where we use some fixed, predefined features that contain topological information \citep[e.g.,][]{dindin2020topological,umeda2017time, liu2016applying}. \citet{guss2018characterizing, rieck2018neural} proposed a complexity measure for neural network architectures based on topological data analysis. \citet{CarlssonG2020} applied topological approaches to deep convolutional networks to understand and improve the computations of the network. \citet{hofer2017deep} first developed a technique to input persistence diagrams into neural networks by introducing their own topological layer. \cite{CarriereCILRU2020} proposed a network layer for persistence diagrams built on top of graphs.  \citet{PoulenardSO2018, toplayerml2019, hofer2019connectivity, MoorHRB2020} also proposed various topology loss functions and layers applied to deep learning. Nevertheless, all the previous approaches suffer from at least one or more of the following limitations: 1) they rely on a particular parametrized map or filtration, 2) they lack stability results or the stability is limited to a particular type of input data representation, and 3) most importantly, the differentiability of persistent homology is not guaranteed with respect to the layer's input therefore we can not place the layer in the middle of deep networks in general.

\begin{figure}
\centering
  \includegraphics[width=.9\linewidth]{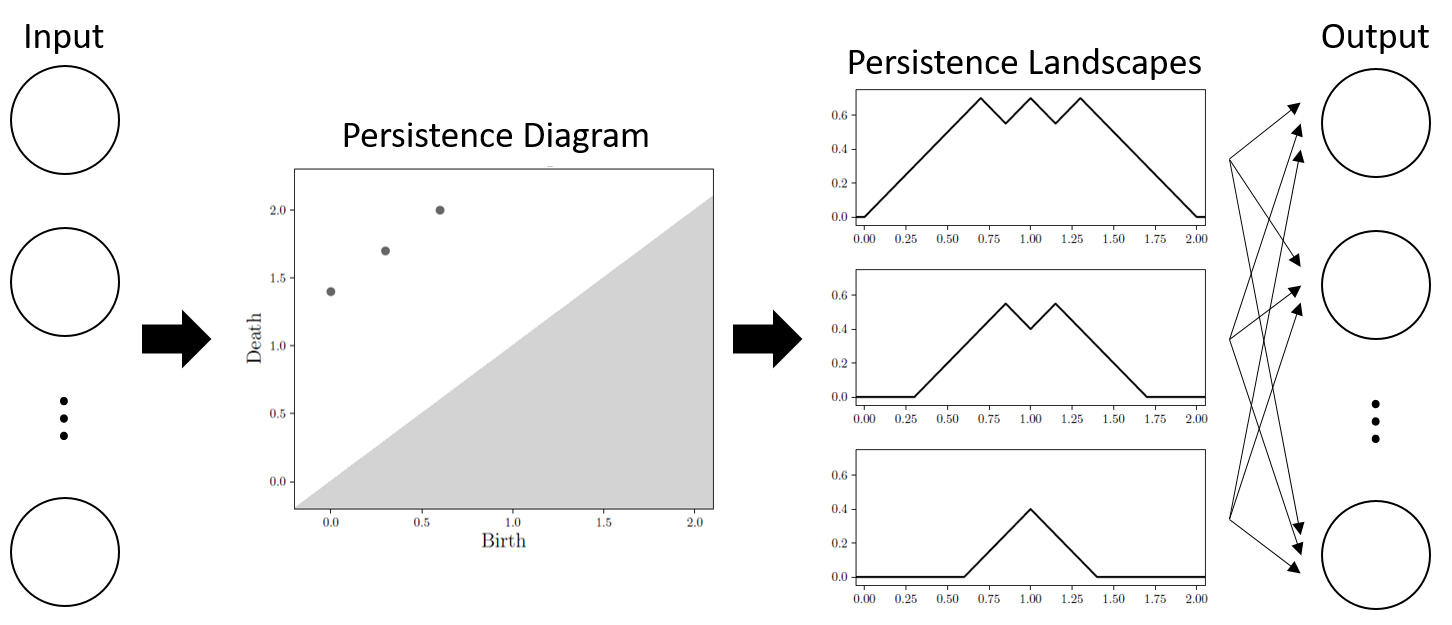}
  \caption{Illustration of \ourtoplayer, a novel topological layer based on weighted persistence landscapes. Information in the persistence diagram is first encoded into persistence landscapes as a form of vectorized function, and then a deep learning model determines which components of the landscape (e.g., particular hills or valleys) are important for a given task during training. \ourtoplayer can be placed anywhere in the  network.
  }
  \label{fig:pllay-schematic}
\end{figure}

\textbf{Contribution.}
This paper presents a new topological layer,
\ourtoplayer (Persistence Landscape-based topological Layer: see Figure \ref{fig:pllay-schematic} for an illustration), that does not suffer from the above limitations. Our topological layer does not rely on a particular filtration or a parametrized mapping but still shows favorable theoretical properties. The proposed layer is designed based on the weighted persistence landscapes to be less prone to 
extreme topological distortions. We provide a tight stability bound that does not depend on the input complexity, and show the stability with respect to input perturbations. We also provide a novel adaptation for the DTM function-based filtration, and analyze the stability property. Importantly, we guarantee the differentiability of our layer with respect to the layer's input. 

\textbf{Reproducibility.}
The code for \ourtoplayer is available at \url{https://github.com/jisuk1/pllay/}.

\section{Background and definitions}
\textit{Topological data analysis} (TDA) is a recent and emerging field of data science that relies on topological tools to infer relevant features for possibly complex data \citep{carlsson2009topology}. In this section, we briefly review basic concepts and main tools in TDA which we will harness to develop our topological layer in this paper. We refer interested readers to \citet{chazal2017introduction, Hatcher2002,EdelsbrunnerH2010,ChazalCGGO2009,ChazalSGO2016} for details and formal definitions. 

 
\subsection{Simplicial complex, persistent homology, and diagrams}
\label{subsec:persistent-homology}

When inferring topological properties of $\mathbb{X}$, a subset of $\mathbb{R}^{d}$, from a finite collection of samples ${X}$, we rely on a  \textit{simplicial complex} $K$, a discrete structure built over the observed points to provide a topological approximation of the underlying space. Two common examples are the \v{C}ech complex and the Vietoris-Rips complex. The \emph{\v{C}ech complex} is the simplicial complex where $k$-simplices correspond to the nonempty intersection of $k+1$ balls centered at vertices. The \emph{Vietoris-Rips} (or simply \emph{Rips}) \emph{complex} is the simplicial complex where simplexes are built based on pairwise distances among its vertices. We refer to Appendix~\ref{sec:simplicial_complex} for formal definitions.
 
A collection of simplicial complexes $\mathcal{F}=\{K_{a}\subset K:a\in\mathbb{R}\}$
satisfying $K_{a}\subset K_{b}$ whenever $a\leq b$ is called a \textit{filtration}
of $K$. A typical way of setting the filtration is through a monotonic
function on the simplex. A function $f\colon K\to\mathbb{R}$ is monotonic
if $f(\varsigma)\leq f(\tau)$ whenever $\varsigma$ is a face of $\tau$
. If we let $K_{a}\coloneqq f^{-1}(-\infty,a]$, then the monotonicity implies
that $K_{a}$ is a subcomplex of $K$ and $K_{a}\subset K_{b}$ whenever
$a\leq b$. In this paper, we assume that the filtration is built
upon a monotonic function.

\textit{Persistent homology} is a multiscale approach to represent the topological features of the complex $K$, and can be represented in the persistence diagram.
For a filtration $\mathcal{F}$ and for each nonnegative $k$, we keep track of when $k$-dimensional homological features (e.g., $0$-dimension: connected component, $1$-dimension: loop, $2$-dimension: cavity,$\ldots$) appear and disappear in the filtration. If a homological feature $\alpha_{i}$ appears at $b_{i}$ and disappears at $d_{i}$, then we say $\alpha_{i}$ is born at $b_{i}$ and dies at $d_{i}$. By considering these pairs $(b_{i},d_{i})$ as points in the plane, one obtains the \textit{persistence diagram} defined as follows.

\begin{definition} \label{def:background_diagram} 	
Let $\mathbb{R}^2_\ast \coloneqq \{ (b,d) \in (\mathbb{R}\cup\infty)^2: d > b \}$. A persistence diagram $\mathcal{D}$ is a finite multiset of $\{p: p \in \mathbb{R}^2_\ast\}$. We let $\mathbb{D}$ denote the set of all such $\mathcal{D}$'s.
\end{definition}

We will use $\mathcal{D}_X, \mathcal{D}_\mathbb{X}$ as shorthand notations for the persistence diagram drawn from the simplicial complex constructed on original data source $X, \mathbb{X}$, respectively.

Lastly, we define the following metrics to measure the distance between two persistence diagrams.
\begin{definition} [Bottleneck and Wasserstein distance]
\label{def:bottleneck-Wasserstein}
 Given two persistence diagrams $\mathcal{D}$ and $\mathcal{D}^{\prime}$, their bottleneck distance ($d_B$) and $q$-th Wasserstein distance ($W_q$) for $q\geq 1$ are defined by
\begin{equation}
d_{B}(\mathcal{D},\mathcal{D}^{\prime})=\inf\limits _{\gamma\in\Gamma}\sup\limits _{p\in\bar{\mathcal{D}}}\|p-\gamma(p)\|_{\infty},\qquad W_{q}(\mathcal{D},\mathcal{D}^{\prime})=\Bigg[\inf\limits _{\gamma\in\Gamma}\sum\limits _{p\in\bar{\mathcal{D}}}\|p-\gamma(p)\|_{\infty}^{q}\Bigg]^{\frac{1}{q}},\label{eq:bottleneck-wasserstein}
\end{equation}
respectively, where $\Vert\cdot\Vert_{\infty}$ is the usual $L_{\infty}$-norm,
$\bar{\mathcal{D}}=\mathcal{D}\cup\text{Diag}$ and $\bar{\mathcal{D}}^{\prime}=\mathcal{D}^{\prime}\cup\text{Diag}$
with $\text{Diag}$ being the diagonal $\{(x,x):x\in\mathbb{R}\}\subset\mathbb{R}^{2}$
with infinite multiplicity, and the set $\Gamma$ consists of all
the bijections $\gamma\colon\bar{\mathcal{D}}\rightarrow\bar{\mathcal{D}}^{\prime}$.
\end{definition}
Note that for all $q \in[1,\infty)$, $d_{B}(\mathcal{D}_X,\mathcal{D}_Y) \leq  W_{q}(\mathcal{D}_X,\mathcal{D}_Y)$ for any given $\mathcal{D}_X, \mathcal{D}_Y$. As $q$ tends to infinity, the Wasserstein distance approaches the bottleneck distance.
Also, see Appendix~\ref{app:ratio-distance} for a further relationship between the bottleneck distance and Wasserstein distance.

\subsection{Persistence landscapes}

A persistence diagram is a multiset, which is difficult to be used as inputs for machine learning methods (due to the complicated space structure, cardinality issues, computationally inefficient metrics, etc.). Hence, it is useful to transform the persistent homology into a functional Hilbert space, where the analysis is easier and learning methods can be directly applied. One good example is the persistence landscape \citep{bubenik2015statistical, bubenik2018persistence, bubenik2017persistence}.
Let $\mathcal{D}$ denote a persistence diagram that contains $N$ off-diagonal birth-death pairs. We first consider a set of piecewise-linear functions $\{\Lambda_p(t)\}_{p\in\mathcal{D}}$ for all birth-death pairs $p = (b,d) \in \mathcal{D}$ as
\[
\Lambda_p(t) = 
\max\{0,\min\{t-b,d-t\}\}.
\]
Then the \textit{persistence landscape} $\lambda$ of the persistence diagram $\mathcal{D}$ is defined as a sequence of functions $\{\lambda_k\}_{k\in \mathbb{N}}$, where
\begin{equation} \label{def:landscape}
\lambda_k(t) = {\kmax}_p \Lambda_p(t), \quad  t\in\mathbb{R}, \ k\in \mathbb{N},
\end{equation}
Hence, the persistence landscape is a set of real-valued functions and is easily computable. Advantages for this kind of functional summaries are discussed in \citet{chazal2014stochastic, berry2018functional}.
		


\subsection{Distance to measure (DTM) function}
\label{subsec:dtm}
The Distance to measure (DTM) \citep{chazal2011geometric, chazal2016rates}
is a robustified version of the distance function. More precisely,
the DTM $d_{\mu,m_{0}}\colon\mathbb{R}^{d}\to\mathbb{R}$ for a probability
distribution $\mu$ with parameter $m_{0}\in(0,1)$ and $r\geq1$
is defined by 
\[
d_{\mu,m_{0}}(x)=\left(\frac{1}{m_{0}}\int_{0}^{m_{0}}(\delta_{\mu,m}(x))^{r}dm\right)^{1/r},
\]
where $\delta_{\mu,m}(x)=\inf\{t>0:\ \mu(\mathbb{B}(x,t))>m\}$ when $\mathbb{B}(x,t)$ is an open ball centered at $x$ with radius $t$. If not specified, $r=2$ is used as a default. In practice, we use a weighted empirical measure
\[
P_{n}(x)=\frac{\sum_{i=1}^{n}\varpi_{i}\mathbbm{1}(X_{i}=x)}{\sum_{i=1}^{n}\varpi_{i}},
\]
with weights $\varpi_i$'s for $\mu$. In this case, we define the \emph{empirical DTM} by
\begin{equation} \label{eq:dtm-empirical}
\hat{d}_{m_{0}}(x) =d_{P_{n},m_{0}}(x)
=\left(\frac{\sum_{X_{i}\in N_{k}(x)}\varpi_{i}'\left\Vert X_{i}-x\right\Vert ^{r}}{m_{0}\sum_{i=1}^{n}\varpi_{i}}\right)^{1/r},
\end{equation}
where $N_{k}(x)$ is the subset of $\{X_{1},\ldots,X_{n}\}$ containing
the $k$ nearest neighbors of $x$, $k$ is such that $\sum_{X_{i}\in N_{k-1}(x)}\varpi_{i}<m_{0}\sum_{i=1}^{n}\varpi_{i}\leq\sum_{X_{i}\in N_{k}(x)}\varpi_{i}$,
and $\varpi_{i}'=\sum_{X_{j}\in N_{k}(x)}\varpi_{j}-m_{0}\sum_{j=1}^{n}\varpi_{j}$
if at least one of $X_{i}$'s is in $N_{k}(x)$ and
$\varpi_{i}'=\varpi_{i}$ otherwise. Hence the empirical DTM behaves
similarly to the $k$-nearest distance with $k=\left\lfloor m_{0}n\right\rfloor $. For i.i.d cases, we typically set $\varpi_{i}=1$ but the weights can be flexibly determined in data-driven way. The parameter $m_0$ determines how much topological/geometrical information should be extracted from the local or global structure. A brief guideline on DTM parameter selection can be found in Appendix~\ref{app:tda-parameter-choice} (see \citet{chazal2011geometric} for more details). Since the resulting persistence diagram is less prone to input perturbations and has nice stability properties, people often prefer using the DTM as their filtration function.

\section{A novel topological layer based on weighted persistence landscapes} \label{sec:algorithm}

In this section, we present a detailed algorithm to implement \ourtoplayer for a general neural network. Let $X$, $\mathcal{D}_X$, $h_{\text{top}}$ denote our input, corresponding persistence diagram induced from $X$, the proposed topological layer, respectively. Broadly speaking, the construction of our proposed topological layer consists of two steps: 1) computing a persistence diagram from the input, and 2) constructing the topological layer from the persistence diagram.

\subsection{Computation of diagram: \texorpdfstring{$X \rightarrow \mathcal{D}_X$}{X->DX}}
\label{sec:compute-diagram}

To compute the persistence diagram from the input data, we first need to define the filtration which requires a simplicial complex $K$ and a function $f\colon K\to\mathbb{R}$. There are several options for $K$ and $f$. We are in general agnostic about which filtration to use since it is in fact problem-dependent; in practice, we suggest using ensemble-like methods that can adapt to various underlying topological structures. One popular choice is the Vietoris-Rips filtration. When there is a one-to-one correspondence between $X_{i}$ and each fixed grid point $Y_{i}$, one obvious choice for $f$ could be just interpreting $X$ as a function values, so $f(Y_{i})=X_{i}$. We refer to \citet{chazal2017introduction} for more examples.


As described in Section~\ref{subsec:dtm}, one appealing choice for $f$ is the DTM function. Due to its favorable properties, the DTM function has been widely used in TDA \citep{AnaiCGIITU2019, XuCGN2019}, and has a good potential for deep learning application. Nonetheless, to the best of our knowledge, the DTM function has not yet been adopted in previous studies. In what follows, we detail two common scenarios for the DTM adaptation: when we consider the input $X$ as 1) data points or 2) weights. 
\begin{table}[h]
\vspace{-3mm}
\begin{multicols}{2}
\begin{itemize}[leftmargin=3mm, itemsep=0mm, partopsep=0pt,parsep=0pt]
\item 
If the input data $X$ is considered as the empirical data points, then the empirical DTM in \eqref{eq:dtm-empirical} with weights
$\varpi_{i}$'s becomes 
\begin{equation} 
\hat{d}_{m_{0}}(x)=\left(\frac{\sum_{X_{i}\in N_{k}(x)}\varpi_{i}'\left\Vert X_{i}-x\right\Vert ^{r}}{m_{0}\sum_{i=1}^{n}\varpi_{i}}\right)^{1/r}, \label{eq:filtration-dtm-location}
\end{equation}
where $k$ and $\varpi_{i}'$ are determined as in \eqref{eq:dtm-empirical}.

\item 
If the input data $X$ is considered as the weights corresponding
to fixed points $\{Y_{1},\ldots,Y_{n}\}$, then the empirical
DTM in \eqref{eq:dtm-empirical} with data points $Y_{i}$'s and weights
$X_{i}$'s becomes 
\begin{equation}
\hat{d}_{m_{0}}(x)=\left(\frac{\sum_{X_{i}\in N_{k}(x)}X_{i}'\left\Vert Y_{i}-x\right\Vert ^{r}}{m_{0}\sum_{i=1}^{n}X_{i}}\right)^{1/r}, 
\label{eq:filtration-dtm-weight}
\end{equation}
where $k$ and $\varpi_{i}'$ are determined as in \eqref{eq:dtm-empirical}.
\end{itemize}
\end{multicols}
\vspace{-10mm}
\end{table}

\begin{figure*}[t!]
\centering
\begin{minipage}[t]{0.58\linewidth}
\centering
\includegraphics[scale=0.26]{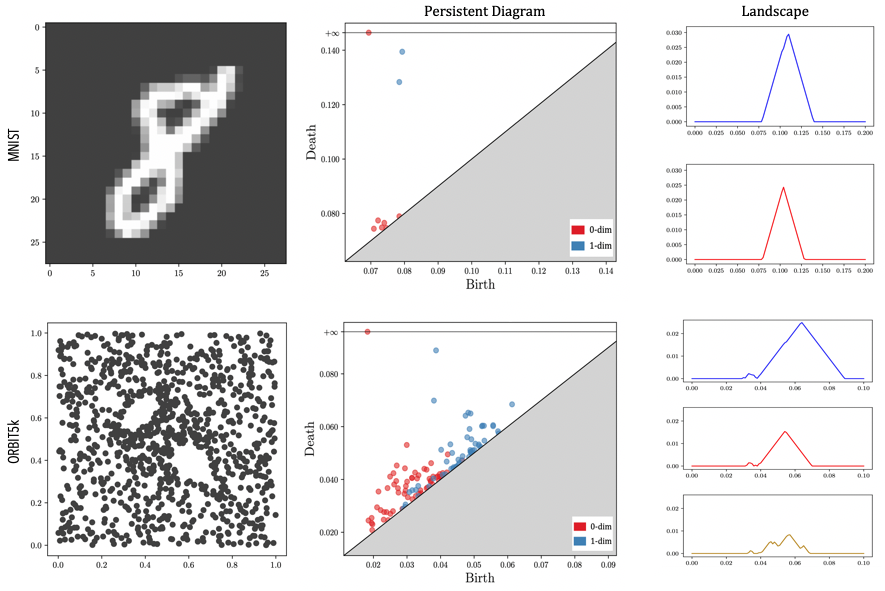}
\captionof{figure}{The topological features encoded in the persistence diagram \& persistence landscapes for MNIST and ORBIT5k sample. In the MNIST example, two loops (1-dimensional feature) in `$8$' are clearly identified and encoded into the 1st and 2nd order landscapes. The ORBIT5k sample shows more involved patterns.}
\label{fig:landscape-example}
\end{minipage}
\hfill%
\begin{minipage}[t]{0.4\linewidth}
\centering
\includegraphics[scale=0.245]{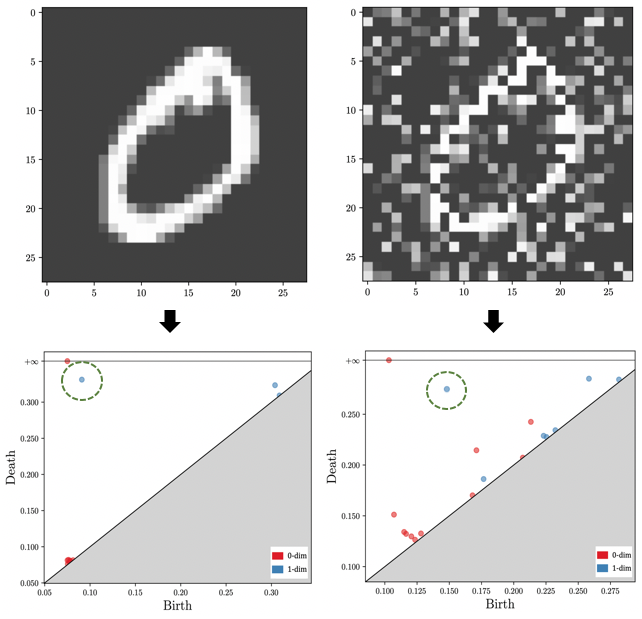}
\captionof{figure}{The significant point (inside green-dashed circle) in the persistence diagram remains almost unchanged even after corrupting pixels and adding noise to the image.}
\label{fig:noise-corruption-example}
\end{minipage}
\end{figure*}

Figure \ref{fig:landscape-example} provides some real data examples (which will be used in Section~\ref{sec:experiments}) of the persistence diagrams and the corresponding persistence landscapes based on the DTM functions. As shown in Figure \ref{fig:noise-corruption-example}, the topological features are expected to be robust to external noise or corruption.

\subsection{Construction of topological layer: \texorpdfstring{$\mathcal{D}_X \rightarrow h_{\text{top}}$}{DX->htop}}
\label{sec:compute-toplayer}

Our topological layer is defined based on a parametrized mapping which takes the persistence diagram $\mathcal{D}$ to be projected onto $\mathbb{R}$, by harnessing persistence landscapes. Our construction is less afflicted by the artificial bending due to a particular transformation procedure as in \citet{hofer2017deep}, yet still guarantees the crucial information in the persistence diagram to be well preserved as will be seen in Section~\ref{sec:stability-theorem}. Insignificant points with low persistence are likely to be ignored systematically without introducing additional nuisance parameters \citep{bubenik2017persistence}.  

Let $\mathbb{R}^{+0}$ denote $[0,\infty)$. Given a persistence diagram $\mathcal{D} \in \mathbb{D}$, we compute the persistence landscape of order $k$ in \eqref{def:landscape}, $\lambda_k(t)$, for $k=1,...,K_{max}$. Then, we compute the weighted average $\overline{\lambda}_{\boldsymbol{\omega}}(t) \coloneqq  \sum_{k=1}^{K_{max}} \omega_k \lambda_k(t)$ with a weight parameter $\boldsymbol{\omega}=\{\omega_k\}_k$, $\omega_k >0, \sum_k \omega_k=1$. Next, we set a domain $[T_{\min},T_{\max}]$ and a resolution $\nu \coloneqq {T}/(m-1)$, and  sample $m$ equal-interval points from $[T_{\min},T_{\max}]$ to obtain $\bm{\overline{\Lambda}_\omega} = \left( \overline{\lambda}_{\boldsymbol{\omega}}(T_{\min}), \overline{\lambda}_{\boldsymbol{\omega}}(T_{\min}+\nu), ... , \overline{\lambda}_{\boldsymbol{\omega}}(T_{\max}) \right)^\top \in \left({\mathbb{R}^{+0}}\right)^m$. Consequently, we have defined a mapping $\bm{\overline{\Lambda}}_{\boldsymbol{\omega}}\colon \mathbb{D} \rightarrow  \left({\mathbb{R}^{+0}}\right)^m$ which is a (vectorized) finite-sample approximation of the weighted persistence landscapes at the resolution $\nu$, at fixed, predetermined locations. Finally, we consider a parametrized differentiable map $g_{\boldsymbol{\theta}}\colon \left({\mathbb{R}^{+0}}\right)^m \rightarrow \mathbb{R}$ which takes the input $\bm{\overline{\Lambda}_\omega}$ and is differentiable with respect to $\boldsymbol{\theta}$ as well. Now, the projection of $\mathcal{D}$ with respect to the mapping $S_{\boldsymbol{\theta},\boldsymbol{\omega}} \coloneqq g_{\boldsymbol{\theta}} \circ \bm{\overline{\Lambda}_\omega}$ defines a single \textit{structure element} for our topological input layer. We summarize the procedure in Algorithm \ref{algorithm:top-layer}. 

\begin{algorithm}[t!]
\caption{Implementation of single structure element for \ourtoplayer} \label{algorithm:top-layer}
\textbf{Input:} persistence diagram $\mathcal{D} \in \mathbb{D}$

\begin{enumerate} 
	\item compute $\lambda_k(t)$ (\ref{def:landscape}) on $t \in [0,T]$ for every $k=1,...,K_{max}$
	\item compute the weighted average $\overline{\lambda}_{\boldsymbol{\omega}}(t) \coloneqq  \sum_{k=1}^{K_{max}} \omega_k \lambda_k(t)$, \ $\omega_k >0, \ \sum_k \omega_k=1$
	\item set $\nu \coloneqq \frac{T}{m-1}$, and compute $\bm{\overline{\Lambda}_\omega} = ( \overline{\lambda}_{\boldsymbol{\omega}}(T_{\min}), \overline{\lambda}_{\boldsymbol{\omega}}(T_{\min}+\nu), ... , \overline{\lambda}_{\boldsymbol{\omega}}(T_{\max}) )^\top \in \mathbb{R}^m$
	\item for a parametrized differentiable map $g_{\boldsymbol{\theta}}\colon \mathbb{R}^m \rightarrow \mathbb{R}$, define $S_{\boldsymbol{\theta},\boldsymbol{\omega}} = g_{\boldsymbol{\theta}} \circ \bm{\overline{\Lambda}_\omega}$ 
\end{enumerate}

\textbf{Output:} $S_{\boldsymbol{\theta},\boldsymbol{\omega}}\colon \mathbb{D} \rightarrow \mathbb{R}$
\end{algorithm}

The projection $S_{\boldsymbol{\theta},\boldsymbol{\omega}}$ is continuous at every $t \in [T_{\min},T_{\max}]$. Also, note that it is differentiable with respect to ${\boldsymbol{\omega}}$ and $\boldsymbol{\theta}$, regardless of the resolution level $\nu$. In what follows, we provide some guidelines that might be useful to implement Algorithm \ref{algorithm:top-layer}.

$\mathbf{{\boldsymbol{\omega}}}$: 
The weight parameter ${\boldsymbol{\omega}}$ can be initialized uniformly, i.e. $\omega_k=1/K_{max}$ for all $k$, and will be re-determined during training through the softmax layer in a way that a certain landscape conveying significant information has more weight. In general, lower-order landscapes tend to be more significant than higher-order landscapes, but the optimal weights may vary from task to task. 

$\mathbf{\boldsymbol{\theta}, g_{\boldsymbol{\theta}}}$:
Likewise, some birth-death pairs, encoded in the landscape function, may contain more crucial information about the topological features of the input data structure than others. Roughly speaking, this is equivalent to say certain mountains (or their ridge or valley) in the landscape are especially important. Hence, the parametrized map $g_{\boldsymbol{\theta}}$ should be able to reflect this by its design. In general, it can be done by affine transformation with scale and translation parameter, followed by an extra nonlinearity and normalization if necessary. We list two possible choices as below.

\begin{itemize}
	\item Affine transformation: with scale and translation parameter $\boldsymbol{\sigma}_i, \boldsymbol{\mu}_i \in \mathbb{R}^m$, $g_{\boldsymbol{\theta_i}}(\bm{\overline{\Lambda}_\omega}) = \boldsymbol{\sigma}_i^\top (\bm{\overline{\Lambda}_\omega} - \boldsymbol{\mu}_i)$ and $\boldsymbol{\theta}_i=(\boldsymbol{\sigma}_i, \boldsymbol{\mu}_i)$.
	\item Logarithmic transformation: with same $\boldsymbol{\theta}_i=({\sigma}_i, \boldsymbol{\mu}_i)$, $g_{\boldsymbol{\theta_i}}(\bm{\overline{\Lambda}_\omega}) = \exp\left(-{\sigma}_i \Vert \bm{\overline{\Lambda}_\omega} - \boldsymbol{\mu}_i \Vert_2 \right)$.
\end{itemize}

Note that other constructions of $g_{\boldsymbol{\theta}}, \boldsymbol{\theta}, {\boldsymbol{\omega}}$ are also possible as long as they satisfy the sufficient conditions described above. Finally, since each structure element corresponds to a single node in a layer, we concatenate many of them, each with different parameters, to form our topological layer.

\begin{definition} [Persistence landscape-based topological layer (\ourtoplayer)]
\label{def:pllay}
	For $n_h \in \mathbb{N}$, let $\boldsymbol{\eta}_i=\left( \boldsymbol{\theta}_i,\boldsymbol{\omega}_i \right)$ denote the set of parameters for the $i$-th structure element and let $\boldsymbol{\eta} = \left(\boldsymbol{\eta}_i \right)_{i=1}^{n_h}$. Given $\mathcal{D}$ and resolution $\nu$, we define {\em \ourtoplayer} as a parametrized mapping with $\boldsymbol{\eta}$ of $\mathbb{D}\rightarrow\mathbb{R}^{n_h}$ such that
	\begin{align} \label{def:top-layer}
		h_{\text{top}}\colon \mathcal{D} \rightarrow \left(S_{\boldsymbol{\eta}_i}(\mathcal{D};\nu) \right)_{i=1}^{n_h}.
	\end{align}
\end{definition} 

Note that this is nothing but a concatenation of $n_h$ topological structure elements (nodes) with different parameter sets (thus $n_h$ is our layer dimension). 

\begin{remark}
Our \emph{\ourtoplayer} considers only $K_{\max}$ top landscape functions. For a given persistence diagram, the points near the diagonal are not likely to appear at $K_{\max}$ top landscape functions, and hence not considered in \emph{\ourtoplayer}. And hence \emph{\ourtoplayer} automatically filters out the noisy features.
\end{remark}

\subsection{Differentiability}
\label{sec:differentiability}
This subsection is devoted to the analysis of the differential behavior of \ourtoplayer with respect to its input (or output from the previous layer), by computing the derivatives $\frac{\partial h_{\text{top}}}{\partial X}$.
Since $\frac{\partial h_{\text{top}}}{\partial X}=\frac{\partial h_{\text{top}}}{\partial\mathcal{D}_{X}}\circ\frac{\partial\mathcal{D}_{X}}{\partial X}$,
this can be done by combining two derivatives $\frac{\partial\mathcal{D}_{X}}{\partial X}$
and $\frac{\partial h_{\text{top}}}{\partial\mathcal{D}_{X}}$. We have extended \citet{PoulenardSO2018} so that we can compute the above derivatives for general persistent homology under arbitrary filtration in our setting. We present the result in Theorem~\ref{thm:derivative-toplayer}.
\begin{thm}

\label{thm:derivative-toplayer}

Let $f$ be the filtration function. Let $\xi$ be a map from each
birth-death point $(b_{i},d_{i})\in\mathcal{D}_{X}$ to a pair of
simplices $(\beta_{i},\delta_{i})$. Suppose that $\xi$ is locally
constant at $X$, and $f(\beta_{i})$ and $f(\delta_{i})$ are differentiable
with respect to $X_{j}$'s.
Then, $h_{\text{top}}$ is differentiable with respect to $X$ and
\begin{align*}
\frac{\partial h_{\text{top}}}{\partial X_{j}} & =\sum_{i}\frac{\partial f(\beta_{i})}{\partial X_{j}}\sum_{l=1}^{m}\frac{\partial g_{\boldsymbol{\theta}}}{\partial x_{l}}\sum_{k=1}^{K_{\max}}\omega_{k}\frac{\partial\lambda_{k}(lv)}{\partial b_{i}} +\sum_{i}\frac{\partial f(\delta_{i})}{\partial X_{j}}\sum_{l=1}^{m}\frac{\partial g_{\boldsymbol{\theta}}}{\partial x_{l}}\sum_{k=1}^{K_{\max}}\omega_{k}\frac{\partial\lambda_{k}(lv)}{\partial d_{i}}.
\end{align*}

\end{thm}

The proof is in Appendix~\ref{proof:derivative-toplayer}. Note that $\frac{\partial\lambda_{k}}{\partial b_{i}}, \frac{\partial\lambda_{k}}{\partial d_{i}}$ are piecewise constant and are easily computed in explicit forms. Also $\frac{\partial g_{\boldsymbol{\theta}}}{\partial x_{l}}$ can be easily realized by an automatic differentiation framework such as \texttt{tensorflow} or \texttt{pytorch}. Our \ourtoplayer in Definition \ref{def:pllay} is thus trainable via backpropagation at an arbitrary location in the network. In Appendix~\ref{app:diffable-dtm}, we also provide a derivative for the DTM filtration.

\section{Stability Analysis}
\label{sec:stability-theorem}

A key property of \ourtoplayer is stability; its discriminating power should remain stable against non-systematic noise or perturbation of input data. 
In this section, we shall provide our theoretical results on the stability properties of the proposed layer. We first address the stability for each structure element with respect to changes in persistence diagrams in Theorem~\ref{thm:stability}.

\begin{thm} \label{thm:stability}
	Let $g_{\boldsymbol{\theta}}$ be $\|\cdot\|_{\infty}$-Lipschitz, i.e. there exists $L_{g}>0$ with $\left\vert g_{\boldsymbol{\theta}}(x)-g_{\boldsymbol{\theta}}(y)\right\vert \leq L_{g}\left\Vert x-y\right\Vert _{\infty}$ for all $x,y \in \mathbb{R}^{m}$. Then for two persistence diagrams $\mathcal{D}, \mathcal{D}^\prime$,
	\begin{align*} 
	 \left\vert	S_{\boldsymbol{\theta},\boldsymbol{\omega}}(\mathcal{D};\nu) - S_{\boldsymbol{\theta},\boldsymbol{\omega}}(\mathcal{D}^\prime;\nu) \right\vert \leq L_{g} d_B(\mathcal{D}, \mathcal{D}^\prime).
	\end{align*}
\end{thm}

Proof of Theorem~\ref{thm:stability} is given in Appendix~\ref{proof:thm-stability}. Theorem~\ref{thm:stability} shows that $S_{\boldsymbol{\theta},\boldsymbol{\omega}}$ is stable with respect to perturbations in the persistence diagram measured by the bottleneck distance \eqref{eq:bottleneck-wasserstein}. It should be noted that only the Lipschitz continuity of $g_{\boldsymbol{\theta}}$ is required to establish the result. 

Next, Corollary~\ref{cor:tightness-claim} shows that under certain conditions our approach improves the previous stability result of \citet{hofer2017deep}.

\begin{cor} \label{cor:tightness-claim}
    For $t>0$, let $n_{t}\in\mathbb{N}$ be satisfying that, 
    for any two diagrams $\mathcal{D}_{t},\mathcal{D}_{t}^{\prime}$ with
    $d_{B}(\mathcal{D},\mathcal{D}_{t})\leq t$ and $d_{B}(\mathcal{D}^{\prime},\mathcal{D}_{t}^{\prime})\leq t$,
    either  $\mathcal{D}_{t}\backslash\mathcal{D}_{t}^{\prime}$
    or $\mathcal{D}_{t}^{\prime}\backslash\mathcal{D}_{t}$ has at least
    $n_{t}$ points. Then, the ratio of our stability bound in Theorem~\ref{thm:stability}
    to that in \cite{hofer2017deep} is upper bounded by 
    \[
    C_{g_{\theta}}/(1+(2t/d_{B}(\mathcal{D},\mathcal{D}^{\prime}))\times(n_{t}-1)),
    \]
    where $C_{g_{\theta}}$ is a constant to be specified in the proof. 
\end{cor}
See Appendix~\ref{proof:tightness-claim} for the proof. Corollary~\ref{cor:tightness-claim} implies that for complex data structures where each $\mathcal{D}$ contains many birth-death pairs (for fixed $t$, in general $n_t$ grows with the increase in the number of points in $\mathcal{D}$), our stability bound is tighter than that of \citet{hofer2017deep} at polynomial rates.

In particular, when we use the DTM function-based filtration proposed in \eqref{eq:filtration-dtm-location}
and \eqref{eq:filtration-dtm-weight}, Theorem~\ref{thm:stability} can be turned into the following stability result with respect to our input
$X$.

\begin{thm}
\label{thm:stability-input}
Suppose $r=2$ is used for the DTM function. Let a differentiable function $g_{\boldsymbol{\theta}}$ and resolution
$\nu$ be given, and let $P$ be a distribution. For the case when
$X_{j}$'s are data points, i.e. when \eqref{eq:filtration-dtm-location}
is used as the DTM function of $X$, let $P_{n}$ be the empirical
distribution defined by $P_{n}=\frac{\sum_{i=1}^{n}\varpi_{i}\delta_{X_{i}}}{\sum_{i=1}^{n}\varpi_{i}}$.
For the case when $X_{j}$'s are weights, i.e. when \eqref{eq:filtration-dtm-weight}
is used as the DTM function of $X$, let $P_{n}$ be the empirical
distribution defined by $P_{n}=\frac{\sum_{i=1}^{n}X_{i}\delta_{Y_{i}}}{\sum_{i=1}^{n}X_{i}}$.
Let $\mathcal{D}_{P}$ be the persistence diagram of the DTM filtration
of $P$, and $\mathcal{D}_{X}$ be the persistence diagram of the
DTM filtration of $X$. Then,
\begin{align*}
 & \left\vert S_{\boldsymbol{\theta},\boldsymbol{\omega}}(\mathcal{D}_{X};\nu)-S_{\boldsymbol{\theta},\boldsymbol{\omega}}(\mathcal{D}_{P};\nu)\right\vert
 \leq L_{g} m_{0}^{-1/2}W_{2}(P_{n},P).
\end{align*}
\end{thm}

The proof is given in Appendix~\ref{proof:stability-input}. Theorem~\ref{thm:stability-input} implies that if the empirical distribution
$P_{n}$ induced from the given input $X$ well approximates the true distribution $P$ with respect to the Wasserstein distance, i.e. having small $W_{2}(P_{n},P)$, then
\ourtoplayer constructed on observed data is close to the one as if we were to know the true distribution $P$. 

Theorem~\ref{thm:stability} and \ref{thm:stability-input} suggest that the topological information embedded in the proposed layer is robust against small noise, data corruption, or outliers. We have also discussed the stability result for the Vietoris-Rips and the \v{C}ech complex in Appendix~\ref{sec:stability-rips-cech}.

\section{Experiments} \label{sec:experiments}
To demonstrate the effectiveness of the proposed approach, we study classification problems on two different datasets: \texttt{MNIST} handwritten digits and \texttt{ORBIT5K}. To fairly showcase the benefits of using our proposed method, we keep our network architecture as simple as possible so that we can focus on the contribution from \ourtoplayer. In the experiments, we aim to explore the benefits of our layer through the following questions: 1) does it make the network more robust and reliable against noise, etc.? and 2) does it improve the overall generalization capability compared to vanilla models? In order to address both of these questions, we first consider the \textit{corruption process}, a certain amount of random omission of pixel values or points from each raw example (so we will have less information), and the \textit{noise process}, a certain amount of random addition of uniformly-distributed noise signals or points to each raw example. An example is given in Figure \ref{fig:noise-corruption-example}. Then we fit a standard multilayer perceptron (MLP) and a convolutional neural network (CNN) with and without the augmentation of \ourtoplayer across various noise and corruption rates given to the raw data, and compare the results. The guideline for choosing the TDA parameters in this experiment is described in Appendix~\ref{app:tda-parameter-choice}. We intentionally use a small number of training data ($\sim$1000) so that the convergence rates could be included in the evaluation criteria. Each simulation is repeated $20$ times. We refer to Appendix~\ref{app:exp-setup} for details about each simulation setup and our model architectures. 

\subsection*{MNIST handwritten digits}

We 
classify handwritten digit images from \texttt{MNIST} dataset. Each digit has
distinctive topological information which can be encoded into the Persistence Landscape as
in Figure~\ref{fig:landscape-example}. 

\textbf{Topological layer.} 
We add two parallel {\ourtoplayer}s in Definition \ref{def:top-layer} at the beginning of MLP and CNN models, based on the empirical DTM function in \eqref{eq:filtration-dtm-weight}, where we define fixed $28 \times 28$ points on grid and use a set of grayscale values $X$ as a weight vector for the fixed points. We used $m_{0}=0.05$ and $m_{0}=0.2$ for each layer, respectively (referred to MLP+P, CNN+P(i), respectively). Particularly for the CNN model, it is likely that the output of the convolutional layers might carry significant information about (smoothed) geometry of the input data shape. So we additionally place another \ourtoplayer after each convolutional layer, directly taking the layer output as 2D-function values and using the sublevel filtration (CNN+P). 

\textbf{Baselines.} 
As our baseline methods, we employ 2-layer vanilla MLP, 2-layer CNN, and the topological signature method by \citet{hofer2017deep} based on the empirical DTM function proposed in \eqref{eq:filtration-dtm-weight} (which we will refer to as SLay). The SLay is augmented at the beginning of MLP and CNN, referred to as MLP+S and CNN+S. See Appendix~\ref{app:mnist-setup} for more details.

\begin{figure*}[t!]
\centering
\begin{minipage}[b]{0.67\linewidth}
\centering
\includegraphics[scale=0.52]{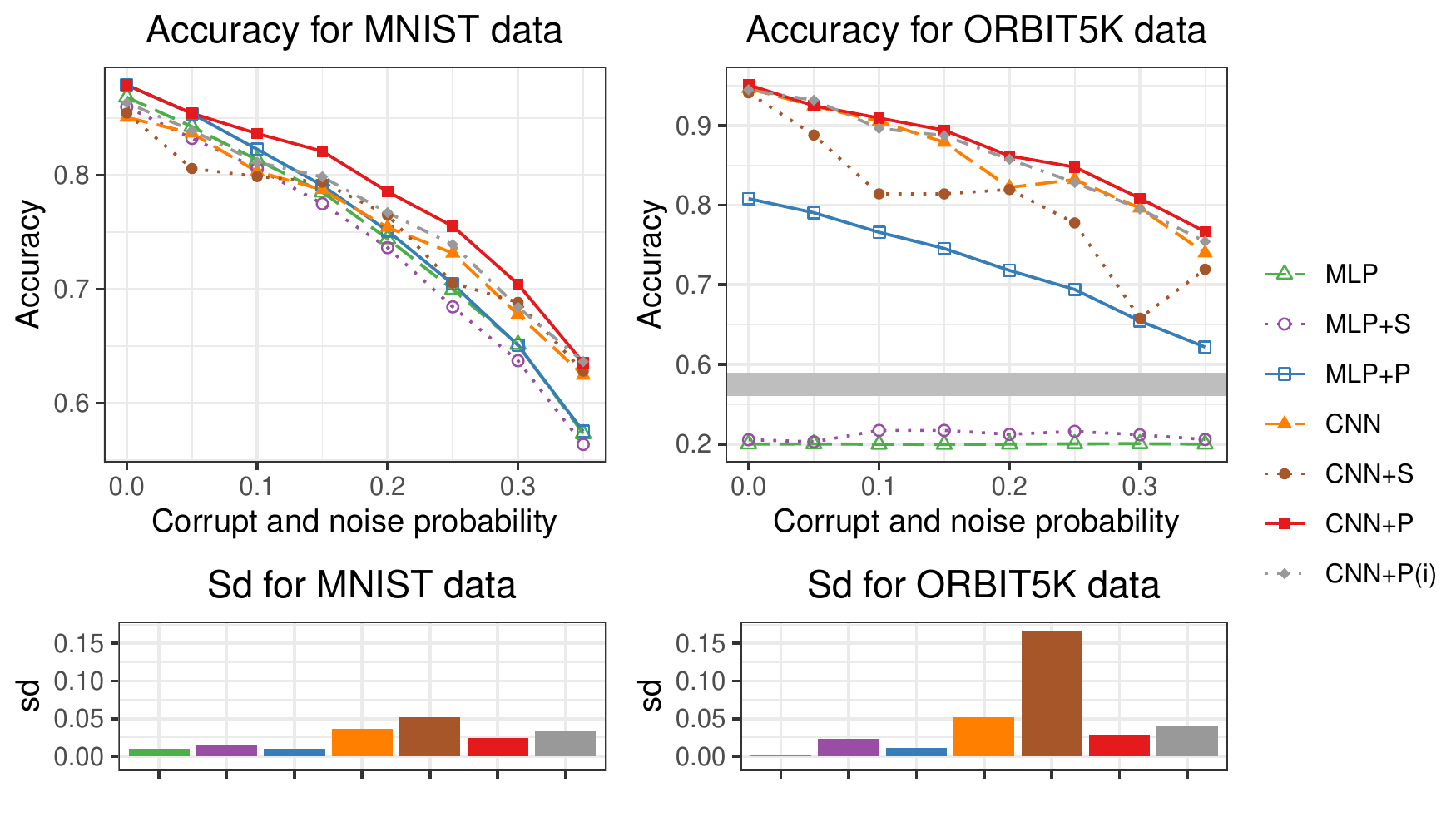}
\captionof{figure}{
Test accuracy in \texttt{MNIST} and \texttt{ORBIT5K} experiments. \ourtoplayer 
consistently improves the accuracy and the robustness against noise and corruption. In particular, in many cases it 
effectively reduces the variance of the classification accuracy on \texttt{ORBIT5K}.
}
\label{fig:result-mnist}
\end{minipage}
\hfill%
\begin{minipage}[b]{0.29\linewidth}
\centering
    \begin{tabular}{c c }
        Model & Accuracy \\ \hline \hline
         PointNet & $0.708$ \\
         & ($\pm 0.285$) \\ \hline
         PersLay & $0.877$ \\
         & ($\pm 0.010$) \\ \hline
         CNN & $0.915$ \\
         & ($\pm 0.088$) \\\hline
         CNN+ & $0.943$ \\ 
         SLay& ($\pm 0.014$) \\ \hline
         \textbf{CNN+}  & \bm{$0.950$}\\
         \textbf{PLLay}& (\bm{$\pm 0.016$})\\
    \end{tabular}
    \captionof{table}{Comparison of different methods for \texttt{ORBIT5K} including the current state-of-the-art PersLay. The proposed method achieves the new state-of-the-art accuracy.}
    \label{tbl:result-3D}
\end{minipage}
\end{figure*}

\textbf{Result.} 
In Figure~\ref{fig:result-mnist}, we observe that \ourtoplayer augmentation consistently improves the accuracy of all the baselines. Interestingly, as we increase the corruption and noise rates, the improvement on CNN increases up to the moderate level of corruption and noise ($\sim 15\%$), then starts to decrease. We conjecture that this is because although DTM filtration is able to robustly capture homological signals as illustrated in Figure 2, if the corruption and noise levels become too much, then the topological structure starts to dissolve in the DTM filtration.

\subsection*{Orbit Recognition}

We classify point clouds generated by $5$ different dynamical systems from \texttt{ORBIT5K} dataset \citep{AdamsEKNPSCHMZ2017, CarriereCILRU2020}. The detailed data generating process is described in Appendix~\ref{app:orbit5k-setup}. 

\textbf{Topological layer.} 
The setup remains the same as in the previous \texttt{MNIST} case, except that 1) \ourtoplayer at the beginning of each network uses the empirical DTM function in \eqref{eq:filtration-dtm-location}, and 2) we set $m_{0}=0.02$.

\textbf{Baselines \& Simulation.} 
All the baseline methods remain the same. For noiseless case, we added PointNet \citep{qi2016pointnet}, a state-of-the-art in point cloud classification, and PersLay \citep{CarriereCILRU2020}, a state-of-the-art in TDA-utilized classification.

\textbf{Result.}
In Figure~\ref{fig:result-mnist}, we observe that \ourtoplayer improves upon MLP and MLP+S by a huge margin ($42\%\sim60\%$). In particular, without augmenting \ourtoplayer, MLP and MLP+S remain at almost a random classifier, which implies that the topological information is indeed crucial for the \texttt{ORBIT5K} classification task, and it would otherwise be very challenging to extract meaningful features. \ourtoplayer improves upon CNN or CNN+S consistently as well. Moreover, it appears that CNN suffers from high variance due to the high complexity of \texttt{ORBIT5K} dataset. On the other hand, \ourtoplayer can effectively mitigate this problem and make the model more stable by utilizing robust topological information from DTM function. Impressively, for the noiseless case,  \ourtoplayer has achieved better performance than all the others including the current state-of-the-art PointNet and PersLay by a large margin.

\section{Discussion}
In this study, we have presented \ourtoplayer, a novel topological layer based on the weighted persistence landscape where we can exploit the topological features effectively. 
We provide the differentiability guarantee of the proposed layer with respect to the layer's input under arbitrary filtration. Hence, our study offers the first general topological layer which can be placed anywhere in the deep learning network. We also present new stability results that verify the robustness and efficiency of our approach. It is worth noting that our method and analytical results in this paper can be extended to silhouettes \citep{chazal2015subsampling, chazal2014stochastic}. In the experiments, we have achieved the new state-of-the-art accuracy for \texttt{ORBIT5K} dataset based on the proposed method. We expect our work to bridge the gap between modern TDA tools and deep learning research.

The computational complexity depends on how \ourtoplayer is used. Computing the DTM is $O(n+m\log n)$ when $m_{0}\propto 1/n$ and k-d tree is used, where $n$ is the input size and $m$ is the grid size. Computing the persistence diagram is $O(m^{2+\epsilon})$ for any small $\epsilon>0$ when the simplicial complex $K$ in Section~\ref{sec:compute-diagram} grows linearly with respect to the grid size such as cubical complex or alpha complex (\cite{ChenK2013} and Theorem 4.4, 5.6 of \cite{BoissonnatCY2018}). Computing the persistence landscape grows linearly with respect to the number of homological features in the persistence diagram, which is the topological complexity of the input and does not necessarily depend on $n$ or $m$. For our experiments, we consider fixed grids of size $28\times28$ and $40\times40$ as in Appendix~\ref{app:exp-setup}, so the computation is not heavy. Also, if we put \ourtoplayer only at the beginning of the deep learning model, then \ourtoplayer can be pre-computed and needs not to be calculated at every epoch in the training.

There are several remarks regarding our experiments. First, we emphasize that SLay in Section~\ref{sec:experiments} is rather an intermediate tool
designed for our simulation and not completely identical to the topological signature method by \citet{hofer2017deep}. For example, SLay combines the method by \citet{hofer2017deep} and the DTM function in \eqref{eq:filtration-dtm-location} and \eqref{eq:filtration-dtm-weight} that have not appeared in the previous study. So we cannot exclude the possibility that the comparable performance of SLay for certain simulations is due to the contribution by the DTM function filtration. Moreover, for CNN, placing extra \ourtoplayer after each convolutional layer appears to bring marginal improvement in accuracy in our experiments. Exploring the optimal architecture with our \ourtoplayer, e.g., finding the most accurate and efficient \ourtoplayer network for a given classification task, would be an interesting future work.

The source code of \ourtoplayer is publicly available at \url{https://github.com/jisuk1/pllay/}.

\section*{Broader Impact}
This paper proposes a novel method of adapting tools in applied mathematics to enhance the learnability of deep learning models. Even though our methodology is generally applicable to any complex modern data, it is not tuned to a specific application that might improperly incur direct societal/ethical consequences. So the broader impact discussion is not needed for our work.

\begin{ack}
During the last 36 months prior to the submission, Jisu Kim received Samsung Scholarship, and Joon Sik Kim received Kwanjeong Fellowship. Fre{\'e}d{\'e}ric Chazal was supported by the ANR AI chair TopAI.
\end{ack}

%% file: toplayer_appendix.tex
\section*{\centerline{APPENDIX}}

\section{Simplicial complex, Persistent homology, and Distance between sets on metric spaces}
\label{sec:simplicial_complex}
Throughout, we will let $\mathbb{X}$ denotes a subset of $\mathbb{R}^{d}$, and ${X}$ denotes a finite collection of points from an arbitrary space $\mathbb{X}$.

A simplicial complex can be seen as a high dimensional generalization
of a graph. Given a set $V$, an \textit{(abstract) simplicial complex}
is a set $K$ of finite subsets of $V$ such that $\alpha\in K$ and
$\beta\subset\alpha$ implies $\beta\in K$. Each set $\alpha\in K$
is called its \textit{simplex}. The \textit{dimension} of a simplex
$\alpha$ is $\dim\alpha=\mathrm{card}\alpha-1$, and the dimension
of the simplicial complex is the maximum dimension of any of its simplices.
Note that a simplicial complex of dimension $1$ is a graph.

When approximating the topology of the underlying space by observed samples, a common choice is the \textit{\v{C}ech complex,} defined next. Below, for any $x\in\mathbb{X}$ and $r>0$, we let $\mathbb{B}_{\mathbb{X}}(x,r)$
denote the open ball centered at $x$ and radius $r>0$ intersected with $\mathbb{X}$. 

\begin{definition}[\v{C}ech complex] \label{def:cech} Let
	$\mathcal{X}\subset\mathbb{X}$ be finite and $r>0$.
	The \emph{(weighted) \v{C}ech complex} is the simplicial complex 
	\begin{equation}
	\textrm{\v{C}ech}^{\mathbb{X}}_{\mathcal{X}}(r)\coloneqq\{ \sigma\subset\mathcal{X}:\ \cap_{x\in\sigma}\mathbb{B}_{\mathbb{X}}(x,r)\neq\emptyset\} .\label{eq:background_cech}
	\end{equation}
	The superscript $\mathbb{X}$ will be dropped when understood from the
	context.
\end{definition}

Another common choice is the \emph{Vietoris-Rips complex}, also referred to as \emph{Rips complex}, where simplexes are built based on pairwise distances among its vertices.

\begin{definition}[Vietoris-Rips complex] \label{def:rips}
    Let	$\mathcal{X}\subset\mathbb{X}$ be finite and $r>0$.	
	The \emph{Vietoris-Rips complex} $\textrm{Rips}_{\mathcal{X}}(r)$ is the simplicial complex defined as
	\begin{equation}
	\textrm{Rips}_{\mathcal{X}}(r)\coloneqq\{ \sigma\subset\mathcal{X}:d(x_{i},x_{j})<2r,\forall x_{i},x_{j}\in\sigma\} .\label{eq:background_rips}
	\end{equation}
\end{definition}

Note that from \eqref{eq:background_cech} and \eqref{eq:background_rips}, the \v{C}ech complex and Vietoris-Rips complex have the following interleaving
inclusion relationship 
\[
\textrm{\v{C}ech}_{\mathcal{X}}(r)\subset \textrm{Rips}_{\mathcal{X}}(r)\subset\textrm{\v{C}ech}_{\mathcal{X}}(2r).
\]
In particular, when $\mathbb{X}\subset\mathbb{R}^{d}$
is a subset of a Euclidean space of dimension $d$, then the constant $2$ can be tightened to $\sqrt{\frac{2d}{d+1}}$ (e.g., see Theorem 2.5 in \cite{deSilvaG2007}):
\[
\textrm{\v{C}ech}_{\mathcal{X}}(r)\subset \textrm{Rips}_{\mathcal{X}}(r)\subset\textrm{\v{C}ech}_{\mathcal{X}}\left(\sqrt{\frac{2d}{d+1}}r\right).
\]


\emph{Persistent homology} \citep{Barannikov1994, zomorodian2005computing, edelsbrunner2000topological, chazal2014persistence} is a multiscale approach to represent topological features of the complex $K$.
A \emph{filtration} $\mathcal{F}$ is a collection of subcomplexes approximating the data points at different resolutions, formally defined as follows. 
\begin{definition}[Filtration]
	A \emph{filtration} $\mathcal{F}=\{ K_{a}\subset K\}_{a\in\mathbb{R}}$
	is a collection of subcomplexes of $K$ such that $a\leq b$ implies that $K_{a}\subset K_{b}$.
	
\end{definition}

For a filtration $\mathcal{F}$ and for each $k\in\mathbb{N}_{0}=\mathbb{N}\cup\{0\}$, the
associated persistent homology $PH_{k}\mathcal{F}$ is an ordered collection of $k$-th dimensional homologies, one for each element of $\mathcal{F}$. 
 
\begin{definition}[Persistent homology] 
	
	Let $\mathcal{F}$ be a filtration and let $k\in\mathbb{N}_{0}$.
	The associated $k$-th \emph{persistent homology} $PH_{k}\mathcal{F}$
	is a collection of groups $\{ H_{k}(K_{a})\} _{a\in\mathbb{R}}$ of each subcomplex $K_{a}$ in $\mathcal{F}$ 
	equipped with homomorphisms $\{ \imath_{k}^{a,b}\} _{a\leq b}$,
	where $H_{k}(K_{a})$ is the $k$-th dimensional homology group of
	$K_{a}$ and $\imath_{k}^{a,b}\colon H_{k}K_{a}\to H_{k}K_{b}$ is the homomorphism induced by the inclusion $K_{a}\subset K_{b}$. 
	
\end{definition}

For the $k$-th persistent homology
$PH_{k}\mathcal{F}$, the set of filtration levels at which a specific
homology appears is always an interval $[b,d)\subset[-\infty,\infty]$,
i.e. a specific homology is formed at some filtration value $b$
and dies when the inside hole is filled at another value $d>b$. To be more formally, the image of a specific homology class $\alpha$ in $H_{k}(K_{a})$ is nonzero if and only if $b\leq a < d$. We often say that $\alpha$ is born at $b$ and dies at $d$. By considering these pairs as points in the plane, one obtains the \emph{persistence diagram} as below.

\begin{definition}[Persistence diagram] 
	Let $\mathbb{R}^2_\ast \coloneqq \{ (b,d) \in (\mathbb{R}\cup\infty)^2: d > b \}$.
	Let $\mathcal{F}$ be a filtration and let $k\in\mathbb{N}_{0}$.
	The corresponding $k$-th persistence diagram $Dgm_{k}(\mathcal{F})$ is a
	finite multiset of $\mathbb{R}^2_\ast$, consisting
	of all pairs $(b,d)$, where $[b,d)$ is the interval of filtration
	values for which a specific homology class appears in $PH_{k}\mathcal{F}$. $b$
	is called a birth time and $d$ is called a death time.
	
\end{definition}

When topological information of the underlying space is approximated
by the observed points, it is often needed to compare two sets with
respect to their metric structures. Here we present two distances
on metric spaces, Hausdorff distance and Gromov-Hausdorff distance.
We refer to \citet{BuragoBI2001} for more details and other distances.

The \emph{Hausdorff distance} \citep[][Definition 7.3.1]{BuragoBI2001} is on sets embedded in the same metric
spaces. This distance measures how two sets are close to each other
in the embedded metric space. When $S\subset\mathbb{X}$, we denote
by $U_{r}(S)$ the $r$-neighborhood of a set $S$ in a metric space,
i.e. $U_{r}(S)=\bigcup_{x\in S}\mathbb{B}_{\mathbb{X}}(x,r)$.

\begin{definition}[Hausdorff distance] 
	
	\label{def:distance_hausdorff}
	
	Let $\mathbb{X}$ be a metric space, and $X,Y\subset\mathbb{X}$
	be a subset. The \emph{Hausdorff distance} between $X$ and $Y$, denoted
	by $d_{H}(X,Y)$, is defined as 
	\[
	d_{H}(X,Y)=\inf\{r>0:\,X\subset U_{r}(Y)\text{ and }Y\subset U_{r}(X)\}.
	\]
	
\end{definition}

The \emph{Gromov-Hausdorff distance} measures how two sets are far
from being isometric to each other. To define the distance, we first
define a relation between two sets called \emph{correspondence}.

\begin{definition}
	
	Let $X$ and $Y$ be two sets. A \emph{correspondence} between $X$
	and $Y$ is a set $C\subset X\times Y$ whose projections to both
	$X$ and $Y$ are both surjective, i.e. for every $x\in X$, there
	exists $y\in Y$ such that $(x,y)\in C$, and for every $y\in Y$,
	there exists $x\in X$ with $(x,y)\in C$.
	
\end{definition}

For a correspondence, we define its \emph{distortion} by how the metric
structures of two sets differ by the correspondence.

\begin{definition}
	
	Let $X$ and $Y$ be two metric spaces, and $C$ be a correspondence
	between $X$ and $Y$. The \emph{distortion} of $C$ is defined by
	\[
	dis(C)=\sup\left\{ \left|d_{X}(x,x')-d_{Y}(y,y')\right|:\,(x,y),(x',y')\in C\right\} .
	\]
	
\end{definition}

Now the Gromov-Hausdorff distance 	\citep[][Theorem 7.3.25]{BuragoBI2001} is defined as the smallest possible
distortion between two sets.

\begin{definition}[Gromov-Hausdorff distance]

	\label{def:distance_gromov_hausdorff}
	
	Let $X$ and $Y$ be two metric spaces. The \emph{Gromov-Hausdorff
		distance} between $X$ and $Y$, denoted as $d_{GH}(X,Y)$, is defined
	as 
	\[
	d_{GH}(X,Y)=\frac{1}{2}\inf_{C}dis(C),
	\]
	where the infimum is over all correspondences between $X$ and $Y$.
	
\end{definition}

\section{Bottleneck distance and Wasserstein distance}
\label{app:ratio-distance}

Our stability bound in Theorem~\ref{thm:stability} is based on the
bottleneck distance, while the stability bound in \citet{hofer2017deep}
is based on Wasserstein distance. Hence to compare these bounds, we
need to understand the relationship between the bottleneck distance
and Wasserstein distance. We already know that the Wasserstein distance is lower bounded by the bottleneck distance. Here, we will find a tighter lower bound for the ratio of the Wasserstein distance to the bottleneck distance.

Before analyzing the relationship between them, we first show a claim. 

\begin{claim}

\label{claim:ratio-distance-cardinality}

Let $\mathcal{D},\mathcal{D}^{\prime}$ be two persistence diagrams.
For $t>0$, let $n_{t}\in\mathbb{N}$ be satisfying the followings:
for any two diagrams $\mathcal{D}_{t},\mathcal{D}_{t}^{\prime}$ with
$d_{B}(\mathcal{D},\mathcal{D}_{t})\leq t$ and $d_{B}(\mathcal{D}^{\prime},\mathcal{D}_{t}^{\prime})\leq t$,
either $\left|\mathcal{D}_{t}\backslash\mathcal{D}_{t}^{\prime}\right|\geq n_{t}$
or $\left|\mathcal{D}_{t}^{\prime}\backslash\mathcal{D}_{t}\right|\geq n_{t}$
holds. Then for any bijection $\gamma:\bar{\mathcal{D}}\to\bar{\mathcal{D}}^{\prime}$,
the number of paired points with being at least $2t$ apart in $L_{\infty}$
distance is greater or equal to $n_{t}$, i.e., 
\[
\left|\left\{ p\in\bar{\mathcal{D}}:\left\Vert p-\gamma(p)\right\Vert _{\infty}>2t\right\} \right|\geq n_{t}.
\]

\end{claim}

And then, we get a lower bound for the ratio of Wasserstein distance
to the bottleneck distance.

\begin{prop}

\label{prop:ratio-distance-wasserstein-bottleneck}

Let $\mathcal{D},\mathcal{D}^{\prime}$ be two persistence diagrams.
For $t>0$, let $n_{t}\in\mathbb{N}$ be satisfying the followings:
for any two diagrams $\mathcal{D}_{t},\mathcal{D}_{t}^{\prime}$ with
$d_{B}(\mathcal{D},\mathcal{D}_{t})\leq t$ and $d_{B}(\mathcal{D}^{\prime},\mathcal{D}_{t}^{\prime})\leq t$,
either $\left|\mathcal{D}_{t}\backslash\mathcal{D}_{t}^{\prime}\right|\geq n_{t}$
or $\left|\mathcal{D}_{t}^{\prime}\backslash\mathcal{D}_{t}\right|\geq n_{t}$
holds. Then, the ratio of $q$-Wasserstein distance to the bottleneck
distance is bounded as 
\[
\frac{W_{q}(\mathcal{D},\mathcal{D}^{\prime})}{d_{B}(\mathcal{D},\mathcal{D}^{\prime})}\geq\left(1+\left(\frac{2t}{d_{B}(\mathcal{D},\mathcal{D}^{\prime})}\right)^{q}(n_{t}-1)\right)^{\frac{1}{q}}.
\]
\end{prop}

\section{Stability for Vietoris-Rips and Cech filtration}
\label{sec:stability-rips-cech}

When we use Vietoris-Rips or \textrm{\v{C}ech} filtration, our result can be turned into the stability result with respect to points in Euclidean space. Let $\mathbb{X}, \mathbb{Y} \subset \mathbb{R}^{d}$ be two bounded sets. 
The next corollary re-states our stability theorem with respect to points in $\mathbb{R}^{d}$.

\begin{cor} \label{cor:stability}
	Let $X,Y$ be any $\epsilon$-coverings of $\mathbb{X}, \mathbb{Y}$, and let $\mathcal{D}_{X}, \mathcal{D}_{Y}$ denote persistence diagrams induced from the Vietoris-Rips or \textrm{\v{C}ech} filtration on $X,Y$ respectively. Then we have
	\begin{equation} 
	\vert	S_{\boldsymbol{\theta}, \boldsymbol{\omega}} (\mathcal{D}_{X};\nu) - S_{\boldsymbol{\theta},\boldsymbol{\omega}}(\mathcal{D}_{Y};\nu) \vert 
	\leq 2 L_{g} \left( d_{GH}(\mathbb{X}, \mathbb{Y}) + 2\epsilon \right).
	\label{eqn:cor-stability}
	\end{equation}		
\end{cor}

The proof is given in Appendix~\ref{proof:cor-stability}. Corollary~\ref{cor:stability} implies that if we assume our observed data points are sufficiently decent quality in the sense that $\epsilon \rightarrow 0$, then our topological layers constructed on those observed points are stable
with respect to small perturbations of the true representation under proper persistent homologies. Here, $\epsilon$ could be interpreted as uncertainty from incomplete sampling. This means the topological information embedded in the proposed layer is robust against small sampling noise or data corruption by missingness.

Moreover, since Gromov-Hausdorff distance is upper bounded by Hausdorff distance, the result in Corollary~\ref{cor:stability} also holds when we use $d_{H}(X, Y)$ in place of $d_{GH}(X, Y)$ in RHS of \eqref{eqn:cor-stability}.

\begin{remark}
In fact, when we have very dense data that have been well-sampled uniformly over the true representation so that $\epsilon \rightarrow 0$, our result in \eqref{eqn:cor-stability} converges to the following:
\[
\left\vert	S_{\boldsymbol{\theta},\boldsymbol{\omega}}(\mathcal{D}_\mathbb{X};\nu) - S_{\boldsymbol{\theta},\boldsymbol{\omega}}(\mathcal{D}_\mathbb{Y};\nu) \right\vert \leq 2 L_{g} 
d_{GH}(\mathbb{X}, \mathbb{Y}).
\]
\end{remark}



\section{Differentiability of DTM function} \label{app:diffable-dtm}
Here we provide a specific example of computing $\frac{\partial f(\varsigma)}{\partial X_{j}}$ when $f$ is the DTM filtration which has not been explored in previous approaches. We first consider the case of \eqref{eq:filtration-dtm-location} where $X_{j}$'s are data points, as in Proposition~\ref{prop:diff-dtm-location}. See Appendix~\ref{proof:diff-dtm-location} for the proof.
\begin{prop}
\label{prop:diff-dtm-location}
When $X_{j}$'s and $\varsigma$ satisfy that $\sum_{X_{i}\in N_{k}(y)}\varpi_{i}\left\Vert X_{i}-y_{l}\right\Vert ^{r}$
are different for each $y_{l}\in\varsigma$, then $f(\varsigma)$
is differentiable with respect to $X_{j}$ and 
\[
\frac{\partial f(\varsigma)}{\partial X_{j}}=\frac{\varpi_{j}'\left\Vert X_{j}-y\right\Vert ^{r-2}(X_{j}-y)I(X_{j}\in N_{k}(y))}{\left(\hat{d}_{m_{0}}(y)\right)^{r-1}m_{0}\sum_{i=1}^{n}\varpi_{i}},
\]
where $I$ is an indicator function and $y=\arg\max_{z\in\varsigma}\hat{d}_{m_{0}}(z)$.
In particular, $f$ is differentiable a.e. with respect to Lebesgue measure on $X$.
\end{prop}

Similarly, we consider the case of \eqref{eq:filtration-dtm-weight} where $X_{j}$'s are weights, as in Proposition~\ref{prop:diff-dtm-weight}. See Appendix~\ref{proof:diff-dtm-weight} for the proof.

\begin{prop}
\label{prop:diff-dtm-weight}
When $X_{j}$'s and $\varsigma$ satisfy that $\sum_{Y_{i}\in N_{k}(y)}X_{i}' \left\Vert Y_{i}-y_{l}\right\Vert ^{r}$
are different for each $y_{l}\in\varsigma$, then $f(\varsigma)$
is differentiable with respect to $X_{j}$ and 
\[
\frac{\partial f(\varsigma)}{\partial X_{j}}=\frac{\left\Vert Y_{j}-y\right\Vert ^{r}I(Y_{j}\in N_{k}(y))-m_{0}\left(\hat{d}_{m_{0}}(y)\right)^{r}}{r\left(\hat{d}_{m_{0}}(y)\right)^{r-1}m_{0}\sum_{i=1}^{n}X_{i}},
\]
where $y=\arg\max_{y\in\varsigma_{i}}\hat{d}_{m_{0}}(y)$.
In particular, $f$ is differentiable a.e. with respect to Lebesgue measure on $X$ and $Y$.
\end{prop}

 Computation of $\frac{\partial h_{\text{top}}}{\partial \boldsymbol{\mu}_i}$, $\frac{\partial h_{\text{top}}}{\partial \boldsymbol{\varsigma}_i}$ are simpler and can be done in a similar fashion. In the experiments, we set $r=2$.

\section{Proofs}

\subsection{Proof of Theorem~\ref{thm:derivative-toplayer}}
\label{proof:derivative-toplayer}

For computing $\frac{\partial h_{\text{top}}}{\partial X_{j}}$, note
that it can be expanded using the chain role as 
\begin{equation}
\frac{\partial h_{\text{top}}}{\partial X_{j}}=\sum_{i}\frac{\partial h_{\text{top}}}{\partial b_{i}}\frac{\partial b_{i}}{\partial X_{j}}+\sum_{i}\frac{\partial h_{\text{top}}}{\partial d_{i}}\frac{\partial d_{i}}{\partial X_{j}},\label{eq:derivative-toplyaer-expand}
\end{equation}
and hence we need to compute $\frac{\partial\mathcal{D}_{X}}{\partial X}=\left\{ \left(\frac{\partial b_{i}}{\partial X_{j}},\frac{\partial d_{i}}{\partial X_{j}}\right)\right\} _{(b_{i},d_{i})\in\mathcal{D}_{X},X_{j}\in X}$
and $\frac{\partial h_{\text{top}}}{\partial\mathcal{D}_{X}}=\left\{ \left(\frac{\partial h_{\text{top}}}{\partial b_{i}},\frac{\partial h_{\text{top}}}{\partial d_{i}}\right)\right\} _{(b_{i},d_{i})\in\mathcal{D}_{X}}$
to compute $\frac{\partial h_{\text{top}}}{\partial X_{j}}$.

We first compute $\frac{\partial\mathcal{D}_{X}}{\partial X}$. Let
$K$ be the simplicial complex, and suppose all the simplices are
ordered in the filtration so that the values of $f$ are nondecreasing,
i.e. if $\varsigma$ comes earlier than $\tau$ then $f(\varsigma)\leq f(\tau)$.
Note that the map $\xi$ from each birth-death point $(b_{i},d_{i})\in\mathcal{D}_{X}$
to a pair of simplices $(\beta_{i},\delta_{i})$ is simply the pairing
returned by the standard persistence diagram \citep{carlsson2005persistence}.
Let $\gamma$ be the homological feature corresponding to $(b_{i},d_{i})$,
then the birth simplex $\beta_{i}$ is the simplex that forms $\gamma$
in $K_{b_{i}}=f^{-1}(-\infty,b_{i}]$, and the death simplex $\delta_{i}$
is the simplex that causes $\gamma$ to collapse in $K_{d_{i}}=f^{-1}(-\infty,d_{i}]$.
For example, if $\gamma$ were to be a $1$-dimensional feature, then
$\beta_{i}$ is the edge in $K_{b_{i}}$ that forms the loop corresponding
to $\gamma$, and $\delta_{i}$ is the triangle in $K_{d_{i}}$ which
incurs the loop corresponding to $\gamma$ can be contracted in $K_{d_{i}}$.

Now, $f(\xi(b_{i}))=f(\beta_{i})=b_{i}$ and $f(\xi(d_{i}))=f(\delta_{i})=d_{i}$,
and from $\xi$ being locally constant on $X$, 
\begin{equation}
\frac{\partial b_{i}}{\partial X_{j}}=\frac{\partial f(\xi(b_{i}))}{\partial X_{j}}=\frac{\partial f(\beta_{i})}{\partial X_{j}},\ \ \frac{\partial d_{i}}{\partial X_{j}}=\frac{\partial f(\xi(d_{i}))}{\partial X_{j}}=\frac{\partial f(\delta_{i})}{\partial X_{j}}.\label{eq:derivative-toplayer-birthdeath}
\end{equation}
Therefore, the derivatives of the birth value and the death value
are the derivatives of the filtration function evaluated at the corresponding
pair of simplices. And $\frac{\partial\mathcal{D}_{X}}{\partial X}=\left\{ \left(\frac{\partial b_{i}}{\partial X_{j}},\frac{\partial d_{i}}{\partial X_{j}}\right)\right\} _{(b_{i},d_{i})\in\mathcal{D}_{X},X_{j}\in X}$
is the collection of these derivatives, hence applying \eqref{eq:derivative-toplayer-birthdeath}
gives 
\begin{equation}
\frac{\partial\mathcal{D}_{X}}{\partial X}=\left\{ \left(\frac{\partial b_{i}}{\partial X_{j}},\frac{\partial d_{i}}{\partial X_{j}}\right)\right\} _{(b_{i},d_{i})\in\mathcal{D}_{X},X_{j}\in X}=\left\{ \left(\frac{\partial f(\beta_{i})}{\partial X_{j}},\frac{\partial f(\delta_{i})}{\partial X_{j}}\right)\right\} _{\xi^{-1}(\beta_{i},\delta_{i})\in\mathcal{D}_{X},X_{j}\in X}.\label{eq:derivative-toplyaer-diagram}
\end{equation}

Now, we compute $\frac{\partial h_{\text{top}}}{\partial\mathcal{D}_{X}}=\left\{ \left(\frac{\partial h_{\text{top}}}{\partial b_{i}},\frac{\partial h_{\text{top}}}{\partial d_{i}}\right)\right\} _{(b_{i},d_{i})\in\mathcal{D}_{X}}$.
Computing $\frac{\partial h_{\text{top}}}{\partial b_{i}}$ can be
done by applying the chain role on $h_{\text{top}}=S_{\boldsymbol{\theta},\boldsymbol{\omega}}=g_{\boldsymbol{\theta}}\circ\bm{\overline{\Lambda}_{\omega}}$
as 
\begin{equation}
\frac{\partial h_{\text{top}}}{\partial b_{i}}=\frac{\partial S_{\boldsymbol{\theta},\boldsymbol{\omega}}}{\partial b_{i}}=\frac{\partial(g_{\boldsymbol{\theta}}\circ\bm{\overline{\Lambda}_{\omega}})}{\partial b_{i}}=\nabla g_{\boldsymbol{\theta}}\circ\frac{\partial\bm{\overline{\Lambda}_{\omega}}}{\partial b_{i}}=\sum_{l=1}^{m}\frac{\partial g_{\boldsymbol{\theta}}}{\partial x_{l}}\frac{\partial\overline{\lambda}_{\boldsymbol{\omega}}(l\nu)}{\partial b_{i}},\label{eq:derivative-toplayer-h-birth-expand}
\end{equation}
where we use $x_{l}$ as the shorthand notation for the input of the
function $g_{\boldsymbol{\theta}}$. Then, applying $\overline{\lambda}_{\boldsymbol{\omega}}(l\nu)=\sum_{k=1}^{K_{max}}\omega_{k}\lambda_{k}(l\nu)$
to \eqref{eq:derivative-toplayer-h-birth-expand} gives 
\begin{equation}
\frac{\partial h_{\text{top}}}{\partial b_{i}}=\sum_{l=1}^{m}\frac{\partial g_{\boldsymbol{\theta}}}{\partial x_{l}}\sum_{k=1}^{K_{\max}}\omega_{k}\frac{\partial\lambda_{k}(lv)}{\partial b_{i}}.\label{eq:derivative-toplayer-h-birth}
\end{equation}
Similarly, $\frac{\partial h_{\text{top}}}{\partial d_{i}}$ can be
computed as 
\begin{equation}
\frac{\partial h_{\text{top}}}{\partial d_{i}}=\sum_{l=1}^{m}\frac{\partial g_{\boldsymbol{\theta}}}{\partial x_{l}}\sum_{k=1}^{K_{\max}}\omega_{k}\frac{\partial\lambda_{k}(lv)}{\partial d_{i}}.\label{eq:derivative-toplayer-h-death}
\end{equation}
And therefore, $\frac{\partial h_{\text{top}}}{\partial\mathcal{D}_{X}}$
is the collection of these derivatives from \eqref{eq:derivative-toplayer-h-birth}
and \eqref{eq:derivative-toplayer-h-death}, i.e., 
\begin{equation}
\frac{\partial h_{\text{top}}}{\partial\mathcal{D}_{X}}=\left\{ \left(\sum_{l=1}^{m}\frac{\partial g_{\boldsymbol{\theta}}}{\partial x_{l}}\sum_{k=1}^{K_{\max}}\omega_{k}\frac{\partial\lambda_{k}(lv)}{\partial b_{i}},\sum_{l=1}^{m}\frac{\partial g_{\boldsymbol{\theta}}}{\partial x_{l}}\sum_{k=1}^{K_{\max}}\omega_{k}\frac{\partial\lambda_{k}(lv)}{\partial d_{i}}\right)\right\} _{(b_{i},d_{i})\in\mathcal{D}_{X}}.\label{eq:derivative-toplayer-h}
\end{equation}

Hence, $\frac{\partial h_{\text{top}}}{\partial X}$ can be computed
by applying \eqref{eq:derivative-toplyaer-diagram} and \eqref{eq:derivative-toplayer-h}
to \eqref{eq:derivative-toplyaer-expand} as 
\begin{align*}
\frac{\partial h_{\text{top}}}{\partial X_{j}} & =\sum_{i}\frac{\partial h_{\text{top}}}{\partial b_{i}}\frac{\partial b_{i}}{\partial X_{j}}+\sum_{i}\frac{\partial h_{\text{top}}}{\partial d_{i}}\frac{\partial d_{i}}{\partial X_{j}}\\
 & =\sum_{i}\frac{\partial f(\beta_{i})}{\partial X_{j}}\sum_{l=1}^{m}\frac{\partial g_{\boldsymbol{\theta}}}{\partial x_{l}}\sum_{k=1}^{K_{\max}}\omega_{k}\frac{\partial\lambda_{k}(lv)}{\partial b_{i}}+\sum_{i}\frac{\partial f(\delta_{i})}{\partial X_{j}}\sum_{l=1}^{m}\frac{\partial g_{\boldsymbol{\theta}}}{\partial x_{l}}\sum_{k=1}^{K_{\max}}\omega_{k}\frac{\partial\lambda_{k}(lv)}{\partial d_{i}}.
\end{align*}

\subsection{Proof of Theorem~\ref{thm:stability}}
\label{proof:thm-stability} 
Let $\mathcal{D}$ and $\mathcal{D}^{\prime}$
be two persistence diagrams and let $\lambda$ and $\lambda^{\prime}$
be their persistence landscapes. All the quantities derived from $\mathcal{D}^{\prime}$
are denoted by a variable name with the superscript $\prime$ hereafter
(e.g., $\lambda_{k}^{\prime}(t),\bm{\overline{\Lambda^{\prime}}}_{\boldsymbol{\omega}}$).

For the stability of the structure element $S_{\boldsymbol{\theta},\boldsymbol{\omega}}$,
we first expand the difference between $S_{\boldsymbol{\theta},\boldsymbol{\omega}}(\mathcal{D};\nu)$
and $S_{\boldsymbol{\theta},\boldsymbol{\omega}}(\mathcal{D}^{\prime};\nu)$
using $S_{\boldsymbol{\theta},\boldsymbol{\omega}}=g_{\boldsymbol{\theta}}\circ\bm{\overline{\Lambda}_{\omega}}$
as 
\begin{equation}
\left\vert S_{\boldsymbol{\theta},\boldsymbol{\omega}}(\mathcal{D};\nu)-S_{\boldsymbol{\theta},\boldsymbol{\omega}}(\mathcal{D}^{\prime};\nu)\right\vert =\left\vert g_{\boldsymbol{\theta}}\left(\bm{\overline{\Lambda}}_{\boldsymbol{\omega}}\right)-g_{\boldsymbol{\theta}}\left(\bm{\overline{\Lambda^{\prime}}}_{\boldsymbol{\omega}}\right)\right\vert .\label{eq:thm-stability-expand}
\end{equation}
Then, RHS of \eqref{eq:thm-stability-expand} is bounded by applying
the Lipschitz condition of the function $g_{\boldsymbol{\theta}}$as
\begin{equation}
\left\vert g_{\boldsymbol{\theta}}\left(\bm{\overline{\Lambda}}_{\boldsymbol{\omega}}\right)-g_{\boldsymbol{\theta}}\left(\bm{\overline{\Lambda^{\prime}}}_{\boldsymbol{\omega}}\right)\right\vert \leq L_{g}\left\Vert \bm{\overline{\Lambda}}_{\boldsymbol{\omega}}-\bm{\overline{\Lambda^{\prime}}}_{\boldsymbol{\omega}}\right\Vert _{\infty}.\label{eq:thm-stability-lipschitz}
\end{equation}
Then for $\left\Vert \bm{\overline{\Lambda}}_{\boldsymbol{\omega}}-\bm{\overline{\Lambda^{\prime}}}_{\boldsymbol{\omega}}\right\Vert _{\infty}$,
note that $\bm{\overline{\Lambda}}_{\boldsymbol{\omega}},\bm{\overline{\Lambda^{\prime}}}_{\boldsymbol{\omega}}\in\mathbb{R}^{m}$,
the $L_{\infty}$ difference of $\bm{\overline{\Lambda}}_{\boldsymbol{\omega}}$
and $\bm{\overline{\Lambda^{\prime}}}_{\boldsymbol{\omega}}$ is bounded
as 
\begin{align}
\left\Vert \bm{\overline{\Lambda}}_{\boldsymbol{\omega}}-\bm{\overline{\Lambda^{\prime}}}_{\boldsymbol{\omega}}\right\Vert _{\infty} & =\max_{0\leq i\leq m-1}\left|\overline{\lambda}_{\boldsymbol{\omega}}(T_{\min}+i\nu)-\overline{\lambda}_{\boldsymbol{\omega}}^{\prime}(T_{\min}+i\nu)\right|\nonumber \\
 & \leq\sup_{t\in[0,T]}\left|\overline{\lambda}_{\boldsymbol{\omega}}(t)-\overline{\lambda}_{\boldsymbol{\omega}}^{\prime}(t)\right|=m^{1/2}\left\Vert \overline{\lambda}_{\boldsymbol{\omega}}-\overline{\lambda}_{\boldsymbol{\omega}}^{\prime}\right\Vert _{\infty}.\label{eq:thm-stability-vector}
\end{align}
Now, for bounding $\left\Vert \overline{\lambda}_{\boldsymbol{\omega}}-\overline{\lambda}_{\boldsymbol{\omega}}^{\prime}\right\Vert _{\infty}$,
we first consider the pointwise difference $\vert\overline{\lambda}_{\boldsymbol{\omega}}(t)-\overline{\lambda}_{\boldsymbol{\omega}}^{\prime}(t)\vert$.
For all $t\in[0,T]$, the difference between $\overline{\lambda}_{\boldsymbol{\omega}}(t)$
and $\overline{\lambda}_{\boldsymbol{\omega}}^{\prime}(t)$ is bounded
as 
\begin{align}
\left|\overline{\lambda}_{\boldsymbol{\omega}}(t)-\overline{\lambda}_{\boldsymbol{\omega}}^{\prime}(t)\right| & =\left\vert \frac{1}{\sum_{k}\omega_{k}}\sum_{k=1}^{K_{\max}}\omega_{k}\lambda_{k}(t)-\frac{1}{\sum_{k}\omega_{k}}\sum_{k=1}^{K_{\max}}\omega_{k}\lambda_{k}^{\prime}(t)\right\vert \nonumber \\
 & \leq\frac{1}{\sum_{k}\omega_{k}}\sum_{k=1}^{K_{\max}}\omega_{k}\left\vert \lambda_{k}(t)-\lambda_{k}^{\prime}(t)\right\vert \nonumber \\
 & \leq\sup_{1\leq k\leq K_{\max},t\in[0,T]}\left|\lambda_{k}(t)-\lambda_{k}^{\prime}(t)\right|=\max_{1\leq k\leq K_{\max}}\left\Vert \lambda_{k}-\lambda_{k}^{\prime}\right\Vert _{\infty}.\label{eq:thm-stability-weighted-average-pointwise}
\end{align}

And hence $\left\Vert \overline{\lambda}_{\boldsymbol{\omega}}-\overline{\lambda}_{\boldsymbol{\omega}}^{\prime}\right\Vert _{\infty}$
is bounded by $\max_{1\leq k\leq K_{\max}}\left\Vert \lambda_{k}-\lambda_{k}^{\prime}\right\Vert _{\infty}$
as well, i.e., 
\begin{equation}
\left\Vert \overline{\lambda}_{\boldsymbol{\omega}}-\overline{\lambda}_{\boldsymbol{\omega}}^{\prime}\right\Vert _{\infty}=\sup_{t\in[0,T]}\left|\overline{\lambda}_{\boldsymbol{\omega}}(t)-\overline{\lambda}_{\boldsymbol{\omega}}^{\prime}(t)\right|\leq\max_{1\leq k\leq K_{\max}}\left\Vert \lambda_{k}-\lambda_{k}^{\prime}\right\Vert _{\infty}.\label{eq:thm-stability-weighted-average}
\end{equation}
Then for all $k=1,\ldots,K_{\max}$, the $\infty$-landscape distance
$\left\Vert \lambda_{k}-\lambda_{k}^{\prime}\right\Vert _{\infty}$
is bounded by the bottleneck distance $d_{B}(\mathcal{D},\mathcal{D}^{\prime})$
from Theorem 13 in \citet{bubenik2015statistical}, i.e. 
\begin{equation}
\left\Vert \lambda_{k}-\lambda_{k}^{\prime}\right\Vert _{\infty}\leq d_{B}(\mathcal{D},\mathcal{D}^{\prime}).\label{eq:thm-stability-landscape}
\end{equation}
Hence, applying \eqref{eq:thm-stability-lipschitz}, \eqref{eq:thm-stability-vector},
\eqref{eq:thm-stability-weighted-average}, \eqref{eq:thm-stability-landscape}
to \eqref{eq:thm-stability-expand} gives the stated stability result
as 
\begin{align*}
\left\vert S_{\boldsymbol{\theta},\boldsymbol{\omega}}(\mathcal{D};\nu)-S_{\boldsymbol{\theta},\boldsymbol{\omega}}(\mathcal{D}^{\prime};\nu)\right\vert  & =\left\vert g_{\boldsymbol{\theta}}\left(\bm{\overline{\Lambda}}_{\boldsymbol{\omega}}\right)-g_{\boldsymbol{\theta}}\left(\bm{\overline{\Lambda^{\prime}}}_{\boldsymbol{\omega}}\right)\right\vert \leq L_{g}\left\Vert \bm{\overline{\Lambda}}_{\boldsymbol{\omega}}-\bm{\overline{\Lambda^{\prime}}}_{\boldsymbol{\omega}}\right\Vert _{\infty}\\
 & \leq L_{g}\left\Vert \overline{\lambda}_{\boldsymbol{\omega}}-\overline{\lambda}_{\boldsymbol{\omega}}^{\prime}\right\Vert _{\infty}\leq L_{g}\max_{1\leq k\leq K_{\max}}\left\Vert \lambda_{k}-\lambda_{k}^{\prime}\right\Vert _{\infty}\\
 & \leq L_{g}d_{B}(\mathcal{D},\mathcal{D}^{\prime}).
\end{align*}

\subsection{Proof of Corollary~\ref{cor:tightness-claim}} 
\label{proof:tightness-claim} 
First note that the result of \citet{hofer2017deep} used $W_{1}$
Wasserstein distance with $L_{r}$ norm for $\forall r\in\mathbb{N}$,
which will be denoted by $W_{1}^{L_{r}}$ in this proof. That is,
\[
W_{1}^{L_{r}}(\mathcal{D},\mathcal{D}^{\prime})\coloneqq\inf\limits _{\gamma}\sum\limits _{p\in\mathcal{D}_{X}}\|p-\gamma(p)\|_{r}
\]
where $\gamma$ ranges over all bijections $\mathcal{D}\rightarrow\mathcal{D}^{\prime}$
(i.e., $W_{1}^{L_{\infty}}$ corresponds to $W_{1}$ in our definition
\ref{def:bottleneck-Wasserstein}). Then, $\left\Vert \cdot\right\Vert _{r}\geq\left\Vert \cdot\right\Vert _{\infty}$
implies that $W_{1}^{L_{r}}$ is lower bounded by $W_{1}$, i.e. 
\begin{equation}
W_{1}^{L_{r}}(\mathcal{D},\mathcal{D}^{\prime})\geq W_{1}(\mathcal{D},\mathcal{D}^{\prime}).\label{eq:tightness-claim-two-wassersteins}
\end{equation}
Now, let $c_{K}$ denote the Lipschitz constant in \citet[][Theorem 1]{hofer2017deep}
and $c_{g_{\theta}}$ denote the constant term in our result in Theorem~\ref{thm:stability},
i.e. $c_{g_{\theta}}=L_{g}\left(\frac{T}{\nu}\right)^{1/2}$. We want
to upper bound the ratio $\frac{c_{g_{\theta}}d_{B}(\mathcal{D},\mathcal{D}^{\prime})}{c_{K}W_{1}^{L_{r}}(\mathcal{D},\mathcal{D}^{\prime})}$.
This directly comes from \eqref{eq:tightness-claim-two-wassersteins}
and Proposition~\ref{prop:ratio-distance-wasserstein-bottleneck}
as 
\[
\frac{c_{g_{\theta}}d_{B}(\mathcal{D},\mathcal{D}^{\prime})}{c_{K}W_{1}^{L_{r}}(\mathcal{D},\mathcal{D}^{\prime})}\geq\frac{c_{g_{\theta}}}{c_{K}}\frac{d_{B}(\mathcal{D},\mathcal{D}^{\prime})}{W_{1}(\mathcal{D},\mathcal{D}^{\prime})}\geq\frac{c_{g_{\theta}}}{c_{K}}\frac{1}{1+\frac{2t}{d_{B}(\mathcal{D},\mathcal{D}^{\prime})}(n_{t}-1)}.
\]
Finally, we define $C_{g_\theta,T,\nu} \coloneqq \frac{c_{g_\theta,T,\nu}}{c_K}$, 
and the result follows.

It should be noted that the bound is actually very loose. However, we can still conclude that our bound is tighter than that of \citet{hofer2017deep} at polynomial rates.

\subsection{Proof of Theorem~\ref{thm:stability-input}}
\label{proof:stability-input}

We first bound the difference between $S_{\boldsymbol{\theta},\boldsymbol{\omega}}(\mathcal{D}_{X};\nu)$
and $S_{\boldsymbol{\theta},\boldsymbol{\omega}}(\mathcal{D}_{P};\nu)$
using Theorem~\ref{thm:stability} as 
\begin{equation}
\left\vert S_{\boldsymbol{\theta},\boldsymbol{\omega}}(\mathcal{D}_{X};\nu)-S_{\boldsymbol{\theta},\boldsymbol{\omega}}(\mathcal{D}_{P};\nu)\right\vert \leq L_{g}d_{B}(\mathcal{D}_{X},\mathcal{D}_{P}).\label{eq:stability-input-toplayer}
\end{equation}
It is left to further bound the bottleneck distance $d_{B}(\mathcal{D}_{X},\mathcal{D}_{P})$.
The bottleneck distance between two diagrams $\mathcal{D}_{X}$ and
$\mathcal{D}_{P}$ is bounded by the stability theorem of persistent
homology as 
\begin{equation}
d_{B}(\mathcal{D}_{X},\mathcal{D}_{P})\leq\left\Vert d_{P_{n},m_{0}}-d_{P,m_{0}}\right\Vert _{\infty}.\label{eq:stability-input-bottleneck}
\end{equation}
Then, from $r=2$ in the DTM function, the $L_{\infty}$ distance
between $d_{P_{n},m_{0}}$ and $d_{P,m_{0}}$ is bounded by the stability
of DTM function (Theorem 3.5 from \citet{chazal2011geometric}) as
\begin{equation}
\left\Vert d_{P_{n},m_{0}}-d_{P,m_{0}}\right\Vert _{\infty}\leq m_{0}^{-1/2}W_{2}(P_{n},P).\label{eq:stability-input-dtm}
\end{equation}
Hence, combining \eqref{eq:stability-input-toplayer}, \eqref{eq:stability-input-bottleneck},
and \eqref{eq:stability-input-dtm} altogether gives the stated stability
result as 
\[
\left\vert S_{\boldsymbol{\theta},\boldsymbol{\omega}}(\mathcal{D}_{X};\nu)-S_{\boldsymbol{\theta},\boldsymbol{\omega}}(\mathcal{D}_{P};\nu)\right\vert \leq L_{g}m_{0}^{-1/2}W_{2}(P_{n},P).
\]

\subsection{Proof of Claim~\ref{claim:ratio-distance-cardinality}}

Let $\gamma\colon\mathcal{D}\to\mathcal{D}^{\prime}$ be any bijection
and let $\mathcal{\mathcal{S}}:=\left\{ p\in\bar{\mathcal{D}}:\left\Vert p-\gamma(p)\right\Vert _{\infty}>2t\right\} $.
Then for $p\in\bar{\mathcal{D}}$ with $\left\Vert p-\gamma(p)\right\Vert _{\infty}\leq2t$,
there exists $\beta(p)\in\mathbb{R}_{*}^{2}$ such that $\left\Vert p-\beta(p)\right\Vert _{\infty}\leq t$
and $\left\Vert \beta(p)-\gamma(p)\right\Vert _{\infty}\leq t$. Now,
define two diagrams $\mathcal{D}_{t},\mathcal{D}_{t}^{\prime}$ as
follows: 
\begin{align*}
\mathcal{D}_{t} & =\mathcal{S}\cup\left\{ \beta(p):p\in\bar{\mathcal{D}}\backslash\mathcal{S}\right\} \backslash Diag,\\
\mathcal{D}_{t}^{\prime} & =\mathcal{S}^{\prime}\cup\left\{ \beta(p):p\in\bar{\mathcal{D}}\backslash\mathcal{S}\right\} \backslash Diag,
\end{align*}
where $\mathcal{S}'\coloneqq\left\{ \gamma(p):p\in\bar{\mathcal{D}}\right\} $.
Then, $d_{B}(\mathcal{D},\mathcal{D}_{t})\leq t$ and $d_{B}(\mathcal{D}^{\prime},\mathcal{D}_{t}^{\prime})\leq t$
from the construction. Hence from the definition of $n_{t}$, either
$\left|\mathcal{D}_{t}\backslash\mathcal{D}_{t}^{\prime}\right|\geq n_{t}$
or $\left|\mathcal{D}_{t}^{\prime}\backslash\mathcal{D}_{t}\right|\geq n_{t}$
holds. Now, note that 
\[
\mathcal{D}_{t}\backslash\mathcal{D}_{t}^{\prime}\subset S\qquad\text{and}\qquad\mathcal{D}_{t}^{\prime}\backslash\mathcal{D}_{t}\subset\mathcal{S}'.
\]
And $\text{\ensuremath{\left|S\right|}=\ensuremath{\left|S^{\prime}\right|}}$,
and hence we get the claimed result as 
\[
\left|S\right|\geq n_{t}.
\]

\subsection{Proof of Proposition~\ref{prop:ratio-distance-wasserstein-bottleneck}}

We consider a bijection $\gamma^{*}$ that realizes the $q$-Wasserstein
distance between $\mathcal{D}$ and $\mathcal{D}^{\prime}$: i.e.
$\gamma^{*}=\underset{\gamma}{\arginf}\sum\limits _{p\in\mathcal{D}}\|p-\gamma(p)\|_{\infty}^{q}$.
Then we have that 
\begin{equation}
d_{B}(\mathcal{D},\mathcal{D}^{\prime})^{q}\leq\sup\limits _{p\in\mathcal{D}}\|p-\gamma^{*}(p)\|_{\infty}^{q}.\label{eqn:appendix-2}
\end{equation}


On the other hand, if we let $p^{*}=\underset{p\in\mathcal{D}}{\argsup}\|p-\gamma^{*}(p)\|_{\infty}$,
we have 
\[
W_{q}(\mathcal{D},\mathcal{D}^{\prime})^{q}=\sum\limits _{p\in\mathcal{D}}\|p-\gamma^{*}(p)\|_{\infty}^{q}=\sup\limits _{p\in\mathcal{D}}\|p-\gamma^{*}(p)\|_{\infty}^{q}+\sum\limits _{p\neq p^{*}}\|p-\gamma^{*}(p)\|_{\infty}^{q}.
\]
Note that from Claim~\ref{claim:ratio-distance-cardinality}, $\left|\left\{ p\in\bar{\mathcal{D}}:\left\Vert p-\gamma^{*}(p)\right\Vert _{\infty}>2t\right\} \right|\geq n_{t}.$
And hence $W_{q}(\mathcal{D},\mathcal{D}^{\prime})^{q}$ can be lower
bounded as 
\begin{align}
W_{q}(\mathcal{D},\mathcal{D}^{\prime})^{q} & =\sup\limits _{p\in\mathcal{D}}\|p-\gamma^{*}(p)\|_{\infty}^{q}+\sum\limits _{p\neq p^{*}}\|p-\gamma^{*}(p)\|_{\infty}^{q}\\
 & \geq\sup\limits _{p\in\mathcal{D}}\|p-\gamma^{*}(p)\|_{\infty}^{q}+(2t)^{q}(n_{t}-1).\label{eq:tightness-claim-wasserstein}
\end{align}
Now, we lower bound the ratio $\frac{W_{q}(\mathcal{D},\mathcal{D}^{\prime})^{q}}{d_{B}(\mathcal{D},\mathcal{D}^{\prime})^{q}}$.
By \eqref{eqn:appendix-2} and \eqref{eq:tightness-claim-wasserstein},
this can be done as follows. 
\begin{align*}
\frac{W_{q}(\mathcal{D},\mathcal{D}^{\prime})^{q}}{d_{B}(\mathcal{D},\mathcal{D}^{\prime})^{q}} & \geq\frac{\sup\limits _{p\in\mathcal{D}}\|p-\gamma^{*}(p)\|_{\infty}^{q}+(2t)^{q}(n_{t}-1)}{d_{B}(\mathcal{D},\mathcal{D}^{\prime})^{q}}\\
 & \geq1+\left(\frac{2t}{d_{B}(\mathcal{D},\mathcal{D}^{\prime})}\right)^{q}(n_{t}-1).
\end{align*}
And hence the ratio of the Wasserstein distance to thw bottleneck
distance $\frac{W_{q}(\mathcal{D},\mathcal{D}^{\prime})}{d_{B}(\mathcal{D},\mathcal{D}^{\prime})}$
is correspondingly lower bounded as 
\[
\frac{W_{q}(\mathcal{D},\mathcal{D}^{\prime})}{d_{B}(\mathcal{D},\mathcal{D}^{\prime})}\geq\left(\frac{W_{q}(\mathcal{D},\mathcal{D}^{\prime})^{q}}{d_{B}(\mathcal{D},\mathcal{D}^{\prime})^{q}}\right)^{\frac{1}{q}}\geq\left(1+\left(\frac{2t}{d_{B}(\mathcal{D},\mathcal{D}^{\prime})}\right)^{q}(n_{t}-1)\right)^{\frac{1}{q}}.
\]

\subsection{Proof of Corollary~\ref{cor:stability}} 
\label{proof:cor-stability} 
The difference between $S_{\boldsymbol{\theta},\boldsymbol{\omega}}(\mathcal{D}_{X};\nu)$
and $S_{\boldsymbol{\theta},\boldsymbol{\omega}}(\mathcal{D}_{Y};\nu)$
is bounded by Theorem~\ref{thm:stability} as 
\begin{equation}
\left\vert S_{\boldsymbol{\theta},\boldsymbol{\omega}}(\mathcal{D}_{X};\nu)-S_{\boldsymbol{\theta},\boldsymbol{\omega}}(\mathcal{D}_{Y};\nu)\right\vert \leq L_{g}d_{B}\left(\mathcal{D}_{X},\mathcal{D}_{Y}\right),\label{eq:cor-stability-toplayer}
\end{equation}
hence it suffices to show 
\begin{equation}
d_{B}\left(\mathcal{D}_{X},\mathcal{D}_{Y}\right)<2\left(d_{GH}\left(\mathbb{X},\mathbb{Y}\right)+2\epsilon\right).\label{eq:cor-stability-sufficient}
\end{equation}
To show \eqref{eq:cor-stability-sufficient}, we first apply the triangle
inequality as 
\begin{align}
d_{B}\left(\mathcal{D}_{X},\mathcal{D}_{Y}\right) & \leq d_{B}\left(\mathcal{D}_{X},\mathcal{D}_{\mathbb{X}}\right)+d_{B}\left(\mathcal{D}_{\mathbb{X}},\mathcal{D}_{\mathbb{Y}}\right)+d_{B}\left(\mathcal{D}_{\mathbb{Y}},\mathcal{D}_{Y}\right).\label{eq:cor-stability-triangle}
\end{align}
And note that since $\mathbb{X},\mathbb{Y},X,Y$ are all bounded in
Euclidean space, they are totally bounded metric spaces. Thus by Theorem
5.2 in \cite{chazal2014persistence}, the bottleneck distance between
any two diagrams is bounded by Gromov-Hausdorff distance, and in particular,
\begin{align}
 & d_{B}\left(\mathcal{D}_{\mathbb{X}},\mathcal{D}_{\mathbb{Y}}\right)\leq2d_{GH}\left(\mathbb{X},\mathbb{Y}\right),\nonumber \\
 & d_{B}\left(\mathcal{D}_{X},\mathcal{D}_{\mathbb{X}}\right)\leq2d_{GH}\left(X,\mathbb{X}\right),\quad d_{B}\left(\mathcal{D}_{\mathbb{Y}},\mathcal{D}_{Y}\right)\leq2d_{GH}\left(\mathbb{Y},Y\right).\label{eq:cor-stability-bottleneck-gh}
\end{align}
And then since the Gromov-Hausdorff distance is bounded by the Hausdorff
distance, 
\begin{equation}
d_{GH}\left(X,\mathbb{X}\right)\leq d_{H}\left(X,\mathbb{X}\right),\quad d_{GH}\left(\mathbb{Y},Y\right)\leq d_{H}\left(\mathbb{Y},Y\right).\label{eq:cor-stability-gh-hausdorff}
\end{equation}
And the Hausdorff distance between $X$ and $\mathbb{X}$ or $Y$
and $\mathbb{Y}$ is bounded by $\epsilon$ by the assumption that
$X,Y$ are $\epsilon$-coverings of $\mathbb{X},\mathbb{Y}$, respectively,
i.e., 
\begin{equation}
d_{H}\left(X,\mathbb{X}\right)<\epsilon,\quad d_{H}\left(\mathbb{Y},Y\right)<\epsilon.\label{eq:cor-stability-covering}
\end{equation}
Hence combining \eqref{eq:cor-stability-triangle}, \eqref{eq:cor-stability-bottleneck-gh},
\eqref{eq:cor-stability-gh-hausdorff}, and \eqref{eq:cor-stability-covering}
gives \eqref{eq:cor-stability-sufficient} as 
\begin{align*}
d_{B}\left(\mathcal{D}_{X},\mathcal{D}_{Y}\right) & \leq d_{B}\left(\mathcal{D}_{X},\mathcal{D}_{\mathbb{X}}\right)+d_{B}\left(\mathcal{D}_{\mathbb{X}},\mathcal{D}_{\mathbb{Y}}\right)+d_{B}\left(\mathcal{D}_{\mathbb{Y}},\mathcal{D}_{Y}\right)\\
 & \leq2\left(d_{GH}\left(X,\mathbb{X}\right)+d_{GH}\left(\mathbb{X},\mathbb{Y}\right)+d_{GH}\left(\mathbb{Y},Y\right)\right)\\
 & \leq2\left(d_{H}\left(X,\mathbb{X}\right)+d_{GH}\left(\mathbb{X},\mathbb{Y}\right)+d_{H}\left(\mathbb{Y},Y\right)\right)\\
 & <2\left(d_{GH}\left(\mathbb{X},\mathbb{Y}\right)+2\epsilon\right).
\end{align*}
Now, the results follows from \eqref{eq:cor-stability-toplayer} and
\eqref{eq:cor-stability-sufficient}.

\subsection{Proof of Proposition~\ref{prop:diff-dtm-location}}
\label{proof:diff-dtm-location}

From \eqref{eq:filtration-dtm-location}, note that for any $y\in\varsigma$,
$\hat{d}_{m_{0}}(y)$ is expanded as 
\begin{equation}
\hat{d}_{m_{0}}(y)=\left(\frac{\sum_{X_{i}\in N_{k}(y)}\varpi_{i}'\left\Vert X_{i}-y\right\Vert ^{r}}{m_{0}\sum_{i=1}^{n}\varpi_{i}}\right)^{1/r},\label{eq:diff-dtm-location-point}
\end{equation}
where $k$ is such that $\sum_{X_{i}\in N_{k-1}(y)}\varpi_{i}<m_{0}\sum_{i=1}^{n}\varpi_{i}\leq\sum_{X_{i}\in N_{k}(y)}\varpi_{i}$,
and $\varpi_{i}'=\sum_{X_{j}\in N_{k}(y)}\varpi_{j}-m_{0}\sum_{j=1}^{n}\varpi_{j}$
for one of $X_{i}$'s that is $k$-th nearest neighbor of $y$ and
$\omega_{i}'=\omega_{i}$ otherwise. Hence, by letting $y=\arg\max_{z\in\varsigma}\hat{d}_{m_{0}}(z)$
applying to \eqref{eq:diff-dtm-location-point}, the filtration function
$f_{X}$ at simplex $\varsigma$ becomes 
\begin{equation}
f_{X}(\varsigma)=\hat{d}_{X,m_{0}}(y)=\left(\frac{\sum_{X_{i}\in N_{k}(y)}\varpi_{i}'\left\Vert X_{i}-y\right\Vert ^{r}}{m_{0}\sum_{i=1}^{n}\varpi_{i}}\right)^{1/r},\label{eq:diff-dtm-location-simplex}
\end{equation}
where the notations $f_{X}$ and $\hat{d}_{X,m_{0}}$ are to clarify
the dependency of $f$ on $X$. And from the condition, $\hat{d}_{m_{0}}(y)>\hat{d}_{m_{0}}(z)$
holds for all $z\in\varsigma$. Hence for sufficiently small $\epsilon>0$
and for any $Z'=\{Z_{1},\ldots,Z_{n}\}$ with $\left\Vert Z_{j}-X_{j}\right\Vert <\epsilon$,
\eqref{eq:diff-dtm-location-simplex} becomes 
\begin{equation}
f_{Z}(\varsigma)=\hat{d}_{Z,m_{0}}(y)=\left(\frac{\sum_{X_{i}\in N_{k}(y)}\varpi_{i}'\left\Vert Z_{i}-y\right\Vert ^{r}}{m_{0}\sum_{i=1}^{n}\varpi_{i}}\right)^{1/r}.\label{eq:diff-dtm-location-simplex-local}
\end{equation}
Hence by differentiating \eqref{eq:diff-dtm-location-simplex-local},
the derivative of $f$ with respect to $X$ is calculated as 
\begin{align*}
\frac{\partial f(\varsigma)}{\partial X_{j}} & =\left(\frac{\sum_{X_{i}\in N_{k}(y)}\varpi_{i}'\left\Vert X_{i}-y\right\Vert ^{r}}{m_{0}\sum_{i=1}^{n}\varpi_{i}}\right)^{\frac{1}{r}-1}\times\frac{\varpi_{j}'\left\Vert X_{j}-y\right\Vert ^{r-2}(X_{j}-y)I(X_{j}\in N_{k}(y))}{m_{0}\sum_{i=1}^{n}\varpi_{i}}\\
 & =\frac{\varpi_{j}'\left\Vert X_{j}-y\right\Vert ^{r-2}(X_{j}-y)I(X_{j}\in N_{k}(y))}{\left(\hat{d}_{m_{0}}(y)\right)^{r-1}m_{0}\sum_{i=1}^{n}\varpi_{i}}.
\end{align*}

\subsection{Proof of Proposition~\ref{prop:diff-dtm-weight}}
\label{proof:diff-dtm-weight}

From \eqref{eq:filtration-dtm-weight}, note that for any $y\in\varsigma$,
$\hat{d}_{m_{0}}(y)$ is expanded as 
\begin{equation}
\hat{d}_{m_{0}}(y)=\left(\frac{\sum_{X_{i}\in N_{k}(y)}X_{i}'\left\Vert Y_{i}-y\right\Vert ^{r}}{m_{0}\sum_{i=1}^{n}X_{i}}\right)^{1/r},\label{eq:diff-dtm-weight-point}
\end{equation}
where $k$ is such that $\sum_{Y_{i}\in N_{k-1}(y)}X_{i}<m_{0}\sum_{i=1}^{n}X_{i}\leq\sum_{Y_{i}\in N_{k}(y)}X_{i}$,
and $X_{i}'=\sum_{X_{j}\in N_{k}(y)}X_{j}-m_{0}\sum_{j=1}^{n}X_{j}$
for one of $Y_{i}$'s that is $k$-th nearest neighbor of $y$ and
$X_{i}'=X_{i}$ otherwise. Hence, by letting $y=\arg\max_{z\in\varsigma}\hat{d}_{m_{0}}(z)$
and applying to \eqref{eq:diff-dtm-weight-point}, the filtration
function $f_{X}$ at simplex $\varsigma$ becomes
\begin{equation}
f_{X}(\varsigma)=\hat{d}_{X,m_{0}}(y)=\left(\frac{\sum_{X_{i}\in N_{k}(y)}X_{i}'\left\Vert Y_{i}-y\right\Vert ^{r}}{m_{0}\sum_{i=1}^{n}X_{i}}\right)^{1/r},\label{eq:diff-dtm-weight-simplex}
\end{equation}
where the notations $f_{X}$ and $\hat{d}_{X,m_{0}}$ are to clarify
the dependency of $f$ on $X$. And from the condition, $\hat{d}_{m_{0}}(y)>\hat{d}_{m_{0}}(z)$
holds for all $z\in\varsigma$. Hence for sufficiently small $\epsilon>0$
and for any $Z'=\{Z_{1},\ldots,Z_{n}\}$ with $\left\Vert Z_{j}-X_{j}\right\Vert <\epsilon$,
\eqref{eq:diff-dtm-weight-simplex} becomes 
\begin{equation}
f_{Z}(\varsigma)=\hat{d}_{Z,m_{0}}(y)=\left(\frac{\sum_{X_{i}\in N_{k}(y)}Z_{i}'\left\Vert Y_{i}-y\right\Vert ^{r}}{m_{0}\sum_{i=1}^{n}Z_{i}}\right)^{1/r}.\label{eq:diff-dtm-weight-simplex-local}
\end{equation}
Hence by differentiating \eqref{eq:diff-dtm-weight-simplex-local},
the derivative of $f$ with respect to $X$ is calculated as 
\begin{align*}
 & \frac{\partial f(\varsigma)}{\partial X_{j}}\\
 & =\frac{1}{r}\left(\frac{\sum_{X_{i}\in N_{k}(y)}X_{i}'\left\Vert Y_{i}-y\right\Vert ^{r}}{m_{0}\sum_{i=1}^{n}X_{i}}\right)^{\frac{1}{r}-1}\times\\
 & \qquad\frac{\left\Vert Y_{j}-y\right\Vert ^{r}I(Y_{j}\in N_{k}(y))\left(m_{0}\sum_{i=1}^{n}X_{i}\right)-m_{0}\left(\sum_{X_{i}\in N_{k}(y)}X_{i}'\left\Vert Y_{i}-y\right\Vert ^{r}\right)}{\left(m_{0}\sum_{i=1}^{n}X_{i}\right)^{2}}\\
 & =\frac{\left\Vert Y_{j}-y\right\Vert ^{r}I(Y_{j}\in N_{k}(y))-m_{0}\left(\hat{d}_{m_{0}}(y)\right)^{r}}{r\left(\hat{d}_{m_{0}}(y)\right)^{r-1}m_{0}\sum_{i=1}^{n}X_{i}}.
\end{align*}

\newpage
\section{Guideline for choosing TDA parameters} \label{app:tda-parameter-choice}
\begin{figure}[t!]
\centering
\begin{subfigure}{0.42\linewidth}\centering\includegraphics[scale=0.57,page=1]{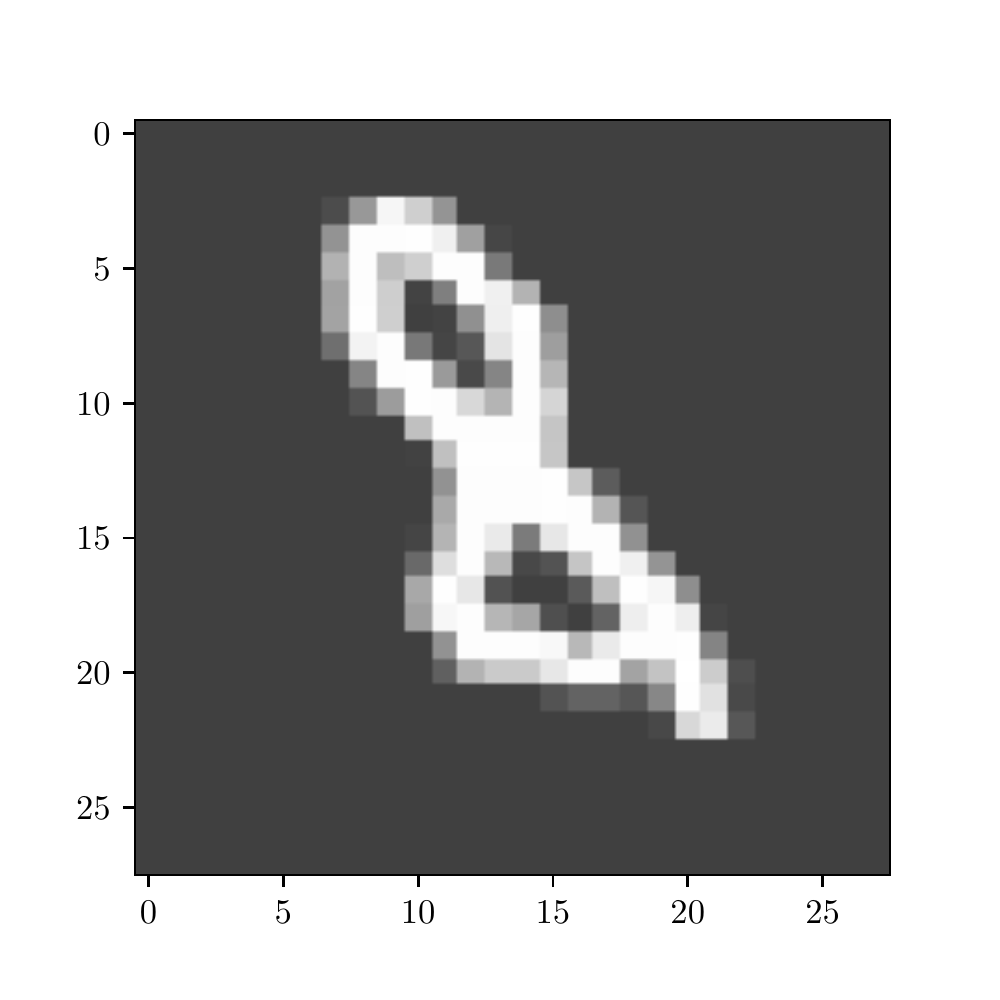}\caption{Digit $8$ in \texttt{MNIST} data.}
\label{subfig:mnist_example_8_tdaparameter_figure}\end{subfigure} 

\begin{subfigure}{0.42\linewidth}\centering\includegraphics[scale=0.57,page=2]{FIG/mnist_example_8_tdaparameter.pdf}\caption{Contour plot of DTM filtration, $m_{0}=0.05$.}
\label{subfig:mnist_example_8_tdaparameter_dtm05}\end{subfigure} \hfill 
\begin{subfigure}{0.56\linewidth}\centering\includegraphics[scale=0.57,page=3]{FIG/mnist_example_8_tdaparameter.pdf}\caption{Persistence Diagram of DTM filtration, $m_{0}=0.05$.} \label{subfig:mnist_example_8_tdaparameter_diagram05}\end{subfigure} 

\begin{subfigure}{0.42\linewidth}\centering\includegraphics[scale=0.57,page=4]{FIG/mnist_example_8_tdaparameter.pdf}\caption{Contour plot of DTM filtration, $m_{0}=0.2$.}
\label{subfig:mnist_example_8_tdaparameter_dtm20}\end{subfigure} \hfill
\begin{subfigure}{0.56\linewidth}\centering\includegraphics[scale=0.57,page=5]{FIG/mnist_example_8_tdaparameter.pdf}\caption{Persistence Diagram of DTM filtration, $m_{0}=0.2$.}
\label{subfig:mnist_example_8_tdaparameter_diagram20}\end{subfigure}

\caption{
One example of the digit $8$ in \texttt{MNIST} data, its contour plots and persistence diagrams of DTM filtration at $m_{0}=0.05$ and $m_{0}=0.2$. When $m_{0}=0.05$, DTM filtration aggregates more locally, and the $1$st persistent homology extracts two loop structures of the digit $8$. When $m_{0}=0.2$, DTM filtration aggregates the digit $8$ more globally, and the $0$th persistent homology extracts three connected component structures of the digit $8$. 
}
\label{fig:mnist_example_8_tdaparameter}
\end{figure}

\ourtoplayer has several TDA parameters to choose: $K_{\max}$, $T_{\min}$, $T_{\max}$, $m$, and $m_0$ if DTM filtration is used. One can try grid search but it could be too time-consuming. More affordable approach is to compute the DTM filtration and the persistence diagram for some data and choose appropriate parameters that can reveal the topological and geometrical information of the data. Figure~\ref{fig:mnist_example_8_tdaparameter} illustrates one example of the digit $8$ in \texttt{MNIST} data. Figure~\ref{fig:mnist_example_8_tdaparameter}(\subref{subfig:mnist_example_8_tdaparameter_figure}) shows the contour plot of the chosen data.

When using a DTM filtration, we need to choose $m_0$ first. DTMs with different $m_0$ values extract different topological and geometrical information. When $m_0$ is small, a DTM filtration aggregates the data more locally, and the geometrical and homological information formed from the local structure is extracted. When $m_0$ is large, a DTM filtration aggregates the data more globally, and the geometrical and homological information formed from the global structure is extracted. From the digit $8$, we would first like to see the two-loop structure. And if we choose $m_{0}=0.05$, then as can be seen in Figure~\ref{fig:mnist_example_8_tdaparameter}(\subref{subfig:mnist_example_8_tdaparameter_dtm05}) and (\subref{subfig:mnist_example_8_tdaparameter_diagram05}), the $1$st persistent homology extracts the two-loop structure, which is more directly expected from the contour plot of the data itself in Figure~\ref{fig:mnist_example_8_tdaparameter}(\subref{subfig:mnist_example_8_tdaparameter_figure}). However, if we choose $m_{0}=0.2$, then as can be seen in Figure~\ref{fig:mnist_example_8_tdaparameter}(\subref{subfig:mnist_example_8_tdaparameter_dtm20}) and (\subref{subfig:mnist_example_8_tdaparameter_diagram20}), the two-loop structure disappears, since the two-loop structure is coming from more local geometry of the data. Meanwhile, as the DTM filtration aggregates the data more globally, the global geometry information that three points on the digit 8(top, center, bottom) being close to neighboring points and being centers of local clusters is extracted in the $0$th persistent homology. For \texttt{MNIST} data, DTM filtrations with $m_{0}=0.05$ and $m_{0}=0.2$ extract different topological and geometrical information of the data. Hence for \texttt{MNIST} data, we used two parallel {\ourtoplayer}s with $m_{0}=0.05$ and $m_{0}=0.2$, respectively.

After choosing $m_0$, choosing other TDA parameters $K_{\max}$, $T_{\min}$, $T_{\max}$, $m$ is more straightforward. One can choose parameters so that the desired topological features are well extracted in the landscape. For $m_{0}=0.05$, as can be seen from Figure~\ref{fig:mnist_example_8_tdaparameter}(\subref{subfig:mnist_example_8_tdaparameter_diagram05}), choosing $K_{\max}=2$, $T_{\min}=0.06$, $T_{\max}=0.3$, $m=25$ will extract two $1$-dimensional features of the persistence diagram in the corresponding landscape. For $m_{0}=0.2$, as can be seen from Figure~\ref{fig:mnist_example_8_tdaparameter}(\subref{subfig:mnist_example_8_tdaparameter_diagram20}), choosing $K_{\max}=3$, $T_{\min}=0.14$, $T_{\max}=0.4$, $m=27$ will extract two $1$-dimensional features of the persistence diagram in the corresponding landscape.

\section{Experiment Details.} \label{app:exp-setup}

 All the experiments were implemented using \texttt{GUDHI} \cite{Gudhi2020} and \texttt{Tensorflow} library in Python and \texttt{TDA} package \cite{FasyKLMMR2014} in R. We use mean and standard deviation across $20$ runs of simulations with different network initializations. We remark that the basic purpose of our experiment design is to highlight the prospects and possibilities of using topological layer, not to win state-of-the-art performances.

\newpage

\subsection{MNIST handwritten digits.} \label{app:mnist-setup}
For MNIST handwritten digits, we use \texttt{MNIST} dataset. Raw input data is a 784 dimensional vector (reshaped from 28 by 28) of real values, each value being the pixel intensity. We use 1000 random samples for the training set and 10000 samples for the test set. Cross-entropy loss was used to train the network for $100$ epochs, using Adam optimizer with mini-batches of size $16$. 

\textbf{Topological layer.} 
For MLP+P and CNN+P(i), we use two parallel {\ourtoplayer}s at the beginning of MLP and CNN models with $32$ nodes each and affine transformation, which are concatenated to the raw input to either MLP or CNN. We used the empirical DTM filtration in \eqref{eq:filtration-dtm-weight}, where we define fixed $28 \times 28$ points on grid on $[-1,1]^{2}$ and use $X$ as a weight vector for the fixed points. For one \ourtoplayer, we used $m_{0}=0.05$, $K_{\max}=2$, $T_{\min}=0.06$, $T_{\max}=0.3$, $m=25$, and for the other \ourtoplayer, we used $m_{0}=0.2$, $K_{\max}=3$, $T_{\min}=0.14$, $T_{\max}=0.4$, $m=27$. For CNN+P, we additionally use one \ourtoplayer after the convolutional layer, with $K_{\max}=3$, $T_{\min}=0.05$, $T_{\max}=0.95$, $m=18$.

\textbf{Baselines.} 
For the baselines, models were designed to have simple structures for quick comparisons:
\begin{itemize} [leftmargin=10pt]
    \item Vanilla MLP: one hidden layer with $64$ units with ReLU activations.
    
    \item CNN: two convolution layers followed by two fully connected layers. 

    \item SLay: for comparison with \ourtoplayer, two SLays are used with 10 nodes each, which are concatenated to the raw input to either MLP or CNN. We used the value $\nu=0.005$ and $\nu=0.01$ for the hyperparameter of each SLay, respectively.
\end{itemize}

\begin{table}[t!]
\begin{center}
\begin{tabular}{|c|c|c|c|c|c|c|c|c|}
\hline 
\multirow{2}{*}{} & \multicolumn{8}{c|}{Corruption and noise probability}\tabularnewline
\cline{2-9} \cline{3-9} \cline{4-9} \cline{5-9} \cline{6-9} \cline{7-9} \cline{8-9} \cline{9-9} 
 & 0.00 & 0.05 & 0.10 & 0.15 & 0.20 & 0.25 & 0.30 & 0.35\tabularnewline
\hline 
\multirow{2}{*}{MLP} & {\small{}0.8683} & {\small{}0.8425} & {\small{}0.8133} & {\small{}0.7850} & {\small{}0.7441} & {\small{}0.6997} & {\small{}0.6514} & {\small{}0.5732}\tabularnewline
 & {\small{}(0.0063)} & {\small{}(0.0061)} & {\small{}(0.0087)} & {\small{}(0.0086)} & {\small{}(0.0098)} & {\small{}(0.0090)} & {\small{}(0.0124)} & {\small{}(0.0155)}\tabularnewline
\hline 
\multirow{2}{*}{MLP+S} & {\small{}0.8597} & {\small{}0.8322} & {\small{}0.8060} & {\small{}0.7749} & {\small{}0.7364} & {\small{}0.6844} & {\small{}0.6372} & {\small{}0.5637}\tabularnewline
 & {\small{}(0.0087)} & {\small{}(0.0086)} & {\small{}(0.0152)} & {\small{}(0.0147)} & {\small{}(0.0177)} & {\small{}(0.0187)} & {\small{}(0.0213)} & {\small{}(0.0161)}\tabularnewline
\hline 
\multirow{2}{*}{MLP+P} & \textbf{\small{}0.8791} & {\small{}0.8538} & {\small{}0.8227} & {\small{}0.7910} & {\small{}0.7511} & {\small{}0.7045} & {\small{}0.6507} & {\small{}0.5753}\tabularnewline
 & \textbf{\small{}(0.0062)} & {\small{}(0.0061)} & {\small{}(0.0103)} & {\small{}(0.0121)} & {\small{}(0.0109)} & {\small{}(0.0087)} & {\small{}(0.0120)} & {\small{}(0.0135)}\tabularnewline
\hline 
\multirow{2}{*}{CNN} & {\small{}0.8506} & {\small{}0.8367} & {\small{}0.8030} & {\small{}0.7872} & {\small{}0.7541} & {\small{}0.7315} & {\small{}0.6778} & {\small{}0.6245}\tabularnewline
 & {\small{}(0.0261)} & {\small{}(0.0246)} & {\small{}(0.0315)} & {\small{}(0.0340)} & {\small{}(0.0319)} & {\small{}(0.0447)} & {\small{}(0.0506)} & {\small{}(0.0478)}\tabularnewline
\hline 
\multirow{2}{*}{CNN+S} & {\small{}0.8544} & {\small{}0.8058} & {\small{}0.7988} & {\small{}0.7938} & {\small{}0.7649} & {\small{}0.7055} & {\small{}0.6884} & {\small{}0.6281}\tabularnewline
 & {\small{}(0.0194)} & {\small{}(0.1081)} & {\small{}(0.0252)} & {\small{}(0.0326)} & {\small{}(0.0215)} & {\small{}(0.1268)} & {\small{}(0.0372)} & {\small{}(0.0407)}\tabularnewline
\hline 
\multirow{2}{*}{CNN+P} & {\small{}0.8790} & \textbf{\small{}0.8541} & \textbf{\small{}0.8364} & \textbf{\small{}0.8209} & \textbf{\small{}0.7855} & \textbf{\small{}0.7551} & \textbf{\small{}0.7044} & {\small{}0.6355}\tabularnewline
 & {\small{}(0.0151)} & \textbf{\small{}(0.0218)} & \textbf{\small{}(0.0214)} & \textbf{\small{}(0.0217)} & \textbf{\small{}(0.0247)} & \textbf{\small{}(0.0289)} & \textbf{\small{}(0.0230)} & {\small{}(0.0404)}\tabularnewline
\hline 
\multirow{2}{*}{CNN+P(i)} & {\small{}0.8635} & {\small{}0.8391} & {\small{}0.8113} & {\small{}0.7985} & {\small{}0.7671} & {\small{}0.7391} & {\small{}0.6841} & \textbf{\small{}0.6364}\tabularnewline
 & {\small{}(0.0189)} & {\small{}(0.0153)} & {\small{}(0.0250)} & {\small{}(0.0275)} & {\small{}(0.0179)} & {\small{}(0.0302)} & {\small{}(0.0936)} & \textbf{\small{}(0.0355)}\tabularnewline
\hline 
\end{tabular}
\end{center}
\caption{Test accuracy in \texttt{MNIST} experiments. In each cell, the top number corresponds
to the average accuracy of the model at the corruption and noise probability,
and the bottom number corresponds to the $1$ standard deviation of
the accuracies. At each column, the model with the best accuracy is
bolded. }
\label{table:expApp_mnist}
\end{table}

\begin{figure}[t!]
\centering
\begin{subfigure}{\linewidth}\centering\includegraphics[scale=0.75,page=1]{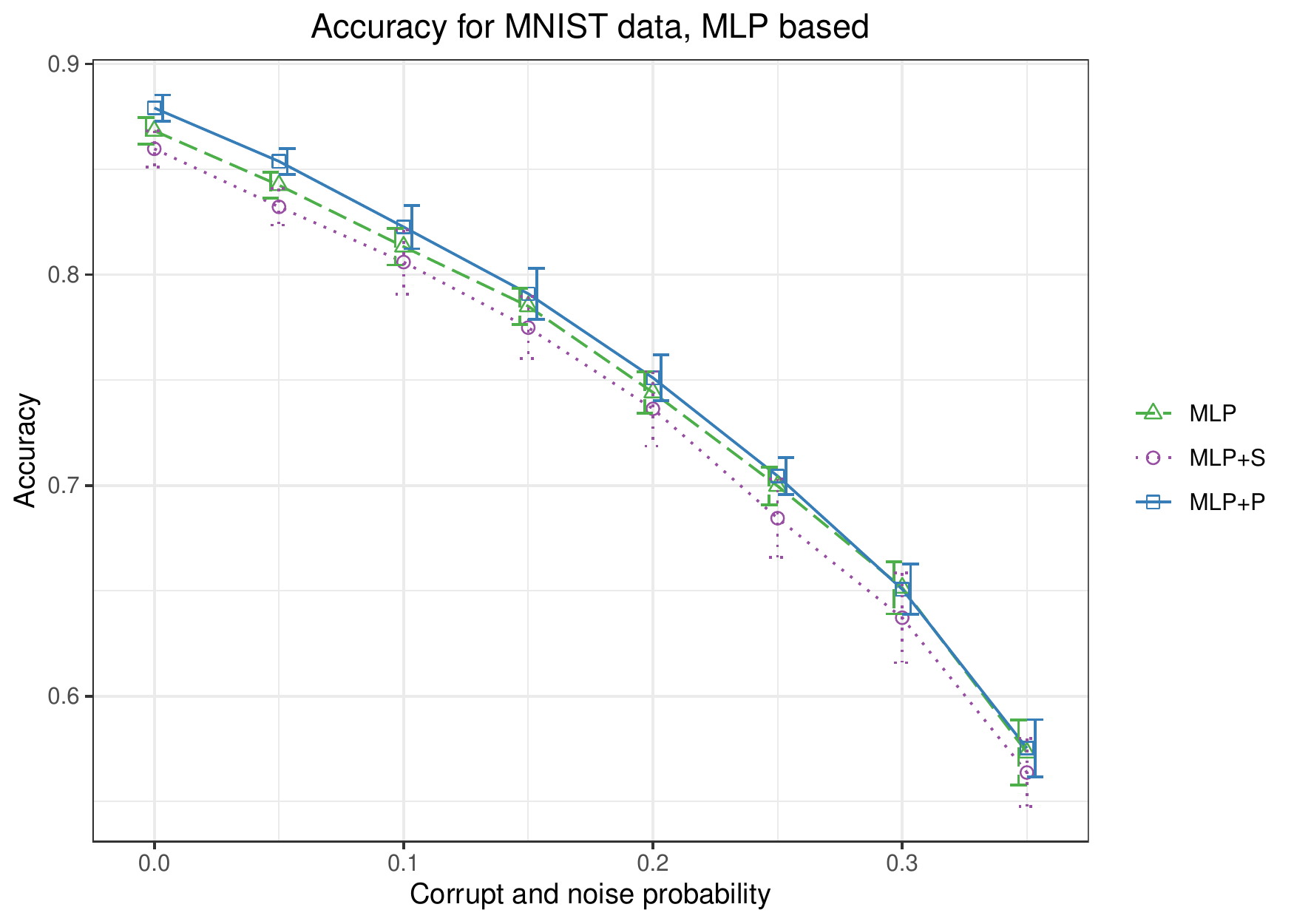}\caption{Test accuracy in \texttt{MNIST} data for MLP, MLP+S, MLP+P.}
\label{subfig:expApp_mnistMlp}\end{subfigure} 

\begin{subfigure}{\linewidth}\centering\includegraphics[scale=0.75,page=2]{FIG/expApp_mnist.pdf}\caption{Test accuracy in \texttt{MNIST} data for CNN, CNN+S, CNN+P, CNN+P(i).}
\label{subfig:expApp_mnistCnn}\end{subfigure}
\caption{
Test accuracy in \texttt{MNIST} experiments. \ourtoplayer contributes to consistent improvement in accuracy and robustness against noise and corruption. In particular, the improvement on CNN increases up to the moderate level of corruption and noise ($\sim 15\%$), and then start to decrease.
}
\label{fig:expApp_mnist}
\end{figure}

\textbf{Result.} 
The Accuracy results for \texttt{MNIST} data in Figure~\ref{fig:result-mnist} is represented with $1$ standard errors in Table~\ref{table:expApp_mnist} and Figure~\ref{fig:expApp_mnist}. In Figure~\ref{fig:expApp_mnist}, the results for MLP, MLP+S, MLP+P are in Figure~\ref{fig:expApp_mnist}(\subref{subfig:expApp_mnistMlp}), and the results for CNN, CNN+S, CNN+P, CNN+P(i) are in Figure~\ref{fig:expApp_mnist}(\subref{subfig:expApp_mnistCnn}). We can see that \ourtoplayer consistently improves the accuracies of all baselines. In particular from Table~\ref{table:expApp_mnist} and Figure~\ref{fig:expApp_mnist}(\subref{subfig:expApp_mnistCnn}), the improvement on CNN is $1.7\%\sim2.8\%$ when the corruption and noise is $0\%\sim 5\%$, and then the improvement goes up to $3.3\%$ when the corruption and noise becomes $10\%\sim15\%$, and then starts to decrease as the corruption and noise further increases. As discussed in Section~\ref{sec:experiments}, this is because although the DTM filtration can robustly capture homological signals up to a moderate amount of corruption and noise, as seen in Figure 2, when the corruption and noise become too much, the topological structure starts to dissolve in the DTM filtration. Also, the accuracies for CNN+P are consistently higher than the accuracies for CNN+P(i), meaning that adding \ourtoplayer in the middle of the network indeed further improves the accuracy.

\newpage

\subsection{Orbit recognition.} \label{app:orbit5k-setup}

For orbit recognition, we use \texttt{ORBIT5K} dataset \citep{AdamsEKNPSCHMZ2017, CarriereCILRU2020}, a synthetic dataset used as a benchmark in Topological Data Analysis. It consists of a point cloud generated by the following discrete dynamical system: given an initial point $(x_{1},y_{1})\in[0,1]^{2}$ and a parameter
$r>0$, we generate a point cloud $\{(x_{n},y_{n})\in[0,1]^{2}:n=1,\ldots,N\}$
as 
\[
\begin{cases}
x_{n+1}=x_{n}+ry_{n}(1-y_{n}) & \text{mod }1,\\
y_{n+1}=y_{n}+rx_{n+1}(1-x_{n+1}) & \text{mod }1.
\end{cases}
\]
For comparison with \citet{AdamsEKNPSCHMZ2017, CarriereCILRU2020}, we use parameters $r=2.5,3.5, 4.0, 4.1, 4.3$, with random initialization of $(x_{1},y_{1})$ and $N=1000$ points in each simulated orbit. We generated $1000$ orbits per each value of $r$, and randomly split the $5000$ observations in $70\%-30\%$ training-test sets as in \citet{CarriereCILRU2020}. Cross-entropy loss was used to train the network for $100$ epochs, using Adam optimizer with mini-batches of size $16$. For the noiseless case, the experiment for PointNet is repeated $5$ times, and the experiment result for PersLay is from \cite{CarriereCILRU2020}.

\textbf{Topological layer.} 
For MLP+P and CNN+P(i), we use one {\ourtoplayer} at the beginning of MLP and CNN models with $64$ nodes and affine transformation, which is solely used as the input to MLP or concatenated to the raw input to CNN. We used the empirical DTM filtration in \eqref{eq:filtration-dtm-location}, where we define fixed $40 \times 40$ points on grid on $[0.0125,0.9875]^{2}$ and use $X$ as the empirical data points. We used $m_{0}=0.01$, $K_{\max}=2$, $T_{\min}=0.03$, $T_{\max}=0.1$, $m=17$. For CNN+P, we additionally use one \ourtoplayer after the convolutional layer, with $K_{\max}=2$, $T_{\min}=0.05$, $T_{\max}=0.95$, $m=18$.

\textbf{Baselines.} 
For the baselines, models were designed to have simple structures for quick comparisons:
\begin{itemize} [leftmargin=10pt]
    \item Vanilla MLP: one hidden layer with $32$ units with ReLU activations.
    
    \item CNN: two convolution layers followed by two fully connected layers. 

    \item SLay: for comparison with \ourtoplayer, one SLay is used with $16$ nodes, which is concatenated to the raw input to either MLP or CNN. We used the value $\nu=0.01$ for the hyperparameter of SLay.
\end{itemize}

\begin{table}[t!]
\begin{center}
\begin{tabular}{|c|c|c|c|c|c|c|c|c|}
\hline 
\multirow{2}{*}{} & \multicolumn{8}{c|}{Noise probability}\tabularnewline
\cline{2-9} \cline{3-9} \cline{4-9} \cline{5-9} \cline{6-9} \cline{7-9} \cline{8-9} \cline{9-9} 
 & 0.00 & 0.05 & 0.10 & 0.15 & 0.20 & 0.25 & 0.30 & 0.35\tabularnewline
\hline 
\multirow{2}{*}{MLP} & {\small{}0.2000} & {\small{}0.2001} & {\small{}0.1997} & {\small{}0.1994} & {\small{}0.1998} & {\small{}0.2003} & {\small{}0.2004} & {\small{}0.1999}\tabularnewline
 & {\small{}(0.0014)} & {\small{}(0.0031)} & {\small{}(0.0020)} & {\small{}(0.0029)} & {\small{}(0.0009)} & {\small{}(0.0010)} & {\small{}(0.0016)} & {\small{}(0.0011)}\tabularnewline
\hline 
\multirow{2}{*}{MLP+S} & {\small{}0.2054} & {\small{}0.2028} & {\small{}0.2171} & {\small{}0.2171} & {\small{}0.2121} & {\small{}0.2159} & {\small{}0.2115} & {\small{}0.2057}\tabularnewline
 & {\small{}(0.0126)} & {\small{}(0.0129)} & {\small{}(0.0364)} & {\small{}(0.0364)} & {\small{}(0.0236)} & {\small{}(0.0301)} & {\small{}(0.0193)} & {\small{}(0.0180)}\tabularnewline
\hline 
\multirow{2}{*}{MLP+P} & {\small{}0.8082} & {\small{}0.7906} & {\small{}0.7660} & {\small{}0.7456} & {\small{}0.7181} & {\small{}0.6942} & {\small{}0.6545} & {\small{}0.6218}\tabularnewline
 & {\small{}(0.0103)} & {\small{}(0.0082)} & {\small{}(0.0115)} & {\small{}(0.0104)} & {\small{}(0.0100)} & {\small{}(0.0130)} & {\small{}(0.0110)} & {\small{}(0.0102)}\tabularnewline
\hline 
\multirow{2}{*}{CNN} & {\small{}0.9466} & {\small{}0.9247} & {\small{}0.9053} & {\small{}0.8791} & {\small{}0.8224} & {\small{}0.8323} & {\small{}0.7963} & {\small{}0.7401}\tabularnewline
 & {\small{}(0.0116)} & {\small{}(0.0152)} & {\small{}(0.0195)} & {\small{}(0.0255)} & {\small{}(0.1474)} & {\small{}(0.0298)} & {\small{}(0.0331)} & {\small{}(0.1293)}\tabularnewline
\hline 
\multirow{2}{*}{CNN+S} & {\small{}0.9412} & {\small{}0.8881} & {\small{}0.8142} & {\small{}0.8142} & {\small{}0.8197} & {\small{}0.7777} & {\small{}0.6580} & {\small{}0.7195}\tabularnewline
 & {\small{}(0.0182)} & {\small{}(0.1612)} & {\small{}(0.1900)} & {\small{}(0.1900)} & {\small{}(0.1473)} & {\small{}(0.1875)} & {\small{}(0.2622)} & {\small{}(0.1778)}\tabularnewline
\hline 
\multirow{2}{*}{CNN+P} & \textbf{\small{}0.9511} & {\small{}0.9249} & \textbf{\small{}0.9095} & \textbf{\small{}0.8941} & \textbf{\small{}0.8619} & \textbf{\small{}0.8480} & \textbf{\small{}0.8087} & \textbf{\small{}0.7668}\tabularnewline
 & \textbf{\small{}(0.0140)} & {\small{}(0.0308)} & \textbf{\small{}(0.0329)} & \textbf{\small{}(0.0305)} & \textbf{\small{}(0.0366)} & \textbf{\small{}(0.0173)} & \textbf{\small{}(0.0396)} & \textbf{\small{}(0.0319)}\tabularnewline
\hline 
\multirow{2}{*}{CNN+P(i)} & {\small{}0.9449} & \textbf{\small{}0.9319} & {\small{}0.8965} & {\small{}0.8873} & {\small{}0.8577} & {\small{}0.8285} & {\small{}0.7954} & {\small{}0.7543}\tabularnewline
 & {\small{}(0.0343)} & \textbf{\small{}(0.0290)} & {\small{}(0.0471)} & {\small{}(0.0143)} & {\small{}(0.0349)} & {\small{}(0.0515)} & {\small{}(0.0516)} & {\small{}(0.0553)}\tabularnewline
\hline 
\end{tabular}
\end{center}
\caption{Test accuracy in \texttt{ORBIT5K} experiments. In each cell, the top number
corresponds to the average accuracy of the model at the noise probability,
and the bottom number corresponds to the $1$ standard deviation of
the accuracies. At each column, the model with the best accuracy is
bolded. }
\label{table:expApp_orbit5k}
\end{table}

\begin{figure}[t!]
\centering
\begin{subfigure}{\linewidth}\centering\includegraphics[scale=0.75,page=1]{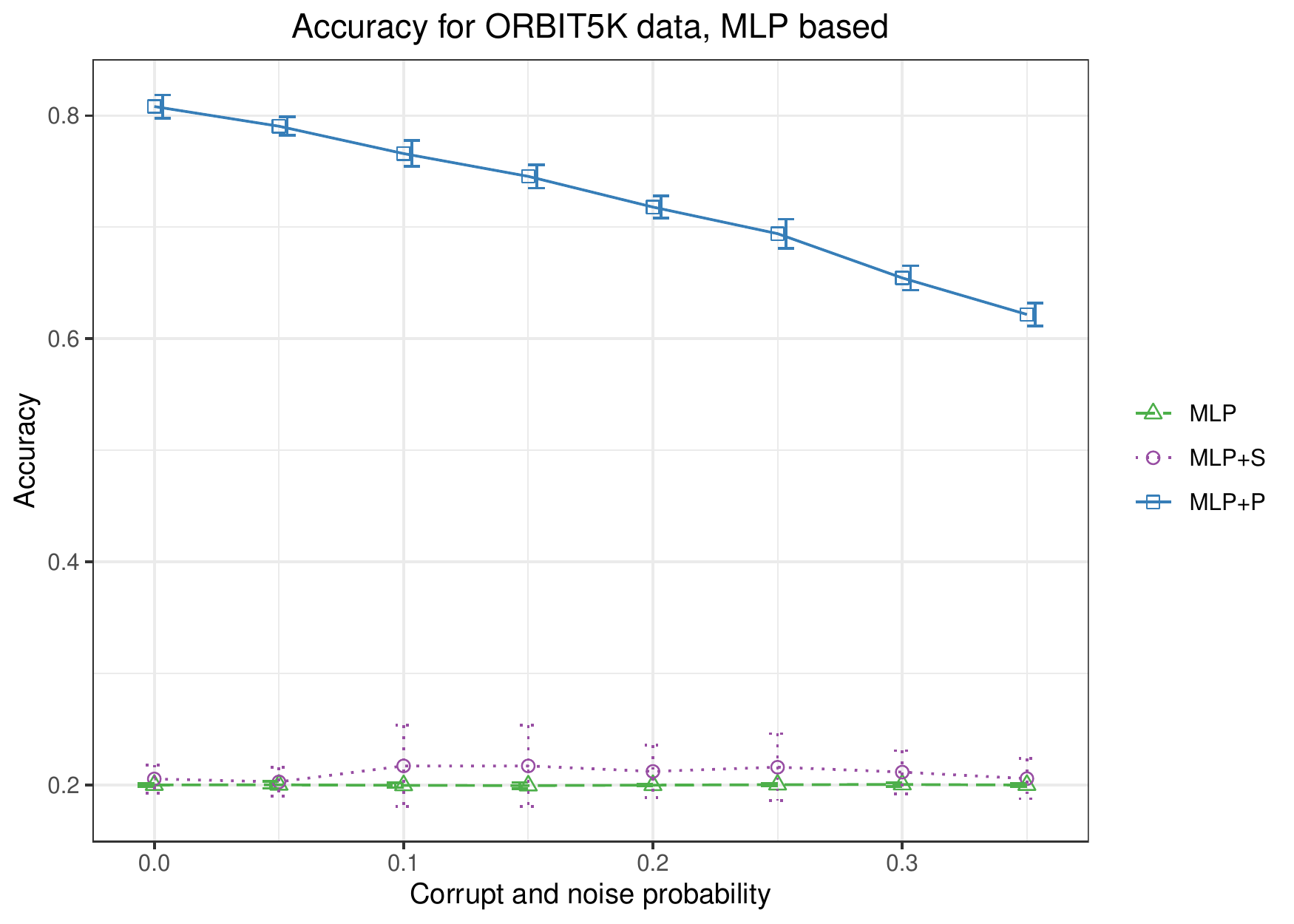}\caption{Test accuracy in \texttt{ORBIT5K} data for MLP, MLP+S, MLP+P.}
\label{subfig:expApp_orbit5kMlp}\end{subfigure} 

\begin{subfigure}{\linewidth}\centering\includegraphics[scale=0.75,page=2]{FIG/expApp_orbit5k.pdf}\caption{Test accuracy in \texttt{ORBIT5K} data for CNN, CNN+S, CNN+P, CNN+P(i).}
\label{subfig:expApp_orbit5kCnn}\end{subfigure}
\caption{
Test accuracy in \texttt{ORBIT5K} experiments. \ourtoplayer contributes to consistent improvement in accuracy and robustness against noise and corruption. In particular in (\subref{subfig:expApp_orbit5kCnn}), when the corruption and noise probability is $0.1, 0.25, 0.35$, \ourtoplayer effectively reduces the variance of classification accuracy.
}
\label{fig:expApp_orbit5k}
\end{figure}

\textbf{Result.} 
The accuracy results for \texttt{ORBIT5K} data in Figure~\ref{fig:result-mnist} is represented with $1$ standard errors in Table~\ref{table:expApp_orbit5k} and Figure~\ref{fig:expApp_orbit5k}. In Figure~\ref{fig:expApp_orbit5k}, the results for MLP, MLP+S, MLP+P are in Figure~\ref{fig:expApp_orbit5k}(\subref{subfig:expApp_orbit5kMlp}), and the results for CNN, CNN+S, CNN+P, CNN+P(i) are in Figure~\ref{fig:expApp_orbit5k}(\subref{subfig:expApp_orbit5kCnn}).
From Figure~\ref{fig:expApp_orbit5k}(\subref{subfig:expApp_orbit5kMlp}), we observe that \ourtoplayer improves over MLP and MLP+S by a huge margin ($42\%\sim60\%$). In particular, without \ourtoplayer, MLP and MLP+S remain at random classifiers, which implies that the topological information is indeed critical for \texttt{ORBIT5K}. In Figure~\ref{fig:expApp_orbit5k}(\subref{subfig:expApp_orbit5kCnn}), \ourtoplayer improves over CNN or CNN+S consistently as well. Moreover, due to the high complexity of \texttt{ORBIT5K}, CNN suffers from high variance at corruption and noise probability $0.2, 0.35$, while \ourtoplayer can effectively reduce the variance at those simulations and make the models more stable by utilizing robust topological information from the DTM function. 
Also, the accuracies for CNN+P are almost always higher than the accuracies for CNN+P(i), meaning that adding \ourtoplayer in the middle of the network indeed further improves the accuracy.

